\documentclass[twoside]{article}

\usepackage[accepted]{aistats2024}

\usepackage[round]{natbib}

\usepackage[T1]{fontenc}
\usepackage[utf8]{inputenc}
\usepackage{tgtermes}

\usepackage{amssymb}
\usepackage{amsthm}
\newcommand{\mathbold}[1]{\ensuremath{\boldsymbol{\mathbf{#1}}}}

\usepackage{scalefnt,letltxmacro}
\LetLtxMacro{\oldtextsc}{\textsc}
\renewcommand{\textsc}[1]{\oldtextsc{\scalefont{1.10}#1}}
\usepackage[ttdefault=true]{AnonymousPro}

\usepackage{amsmath}
\usepackage{dsfont}

\usepackage[english]{babel}
\usepackage[parfill]{parskip}
\usepackage{afterpage}
\usepackage{enumitem}
\usepackage{framed}
\usepackage{xspace}
\usepackage[autostyle]{csquotes}

\usepackage{lineno}

\usepackage{ragged2e}

\newcommand{\parnum}{\bfseries\P\arabic{parcount}}
\newcounter{parcount}
\newcommand\p{%
    \stepcounter{parcount}%
    \leavevmode\marginpar[\hfill\parnum]{\parnum}%
}

\usepackage[normalem]{ulem} %

\usepackage[usenames,dvipsnames]{xcolor}
\definecolor{shadecolor}{gray}{0.9}

\usepackage{graphicx}
\usepackage[labelfont=bf]{caption}
\usepackage[format=hang]{subcaption}
\usepackage{wrapfig}

\usepackage{booktabs, array}
\usepackage{multirow}

\usepackage[algoruled]{algorithm2e}
\SetCommentSty{mycommfont}
\usepackage{listings}
\usepackage{fancyvrb}
\fvset{fontsize=\small}

\usepackage{algorithmic}

\usepackage[colorlinks,linktoc=all]{hyperref}
\usepackage[all]{hypcap}
\hypersetup{citecolor=MidnightBlue}
\hypersetup{linkcolor=MidnightBlue}
\hypersetup{urlcolor=MidnightBlue}
\usepackage[nameinlink,capitalize,noabbrev]{cleveref} %
\creflabelformat{equation}{#2\textup{#1}#3}  %

\usepackage
[acronym,nowarn,section,nogroupskip,nonumberlist]{glossaries}
\glsdisablehyper{}

\lstdefinestyle{mystyle}{
    commentstyle=\color{OliveGreen},
    numberstyle=\tiny\color{black!60},
    stringstyle=\color{BrickRed},
    basicstyle=\ttfamily\scriptsize,
    breakatwhitespace=false,
    breaklines=true,
    captionpos=b,
    keepspaces=true,
    numbers=none,
    numbersep=5pt,
    showspaces=false,
    showstringspaces=false,
    showtabs=false,
    tabsize=2
}
\usepackage{todonotes}
\usepackage{adjustbox}

\DeclareRobustCommand{\KL}[2]{\ensuremath{\textrm{KL}\left(#1\;\|\;#2\right)}}

\DeclareRobustCommand{\Ep}[2]{\ensuremath{\mathds{E}_{#1}\left[#2\right]}}
\DeclareRobustCommand{\E}[1]{\ensuremath{\mathds{E}\left[#1\right]}}

\newcommand{\Ber}{\mathcal{B}\textit{er}}

\newcommand{\Norm}{\mathcal{N}}

\newcommand{\Cat}{\mathcal{C}\textit{at}}

\newcommand{\Unif}{\mathcal{U}}

\newcommand{\veps}{\varepsilon}
\renewcommand{\d}[1]{\ensuremath{\operatorname{d}\!{#1}}}

\DeclareMathOperator*{\argmin}{arg\,min}
\def\indep{\protect\mathpalette{\protect\independenT}{\perp}}
\def\independenT#1#2{\mathrel{\rlap{$#1#2$}\mkern2mu{#1#2}}}
\newcommand{\deq}{\triangleq}  %
\newcommand{\1}{\mathds{1}}  %

\DeclareRobustCommand{\sd}[1]{\color{black!80!white}\scriptstyle #1}

\newcommand{\mba}{\mathbold{a}}
\newcommand{\mbb}{\mathbold{b}}
\newcommand{\mbc}{\mathbold{c}}

\newcommand{\mbm}{\mathbold{m}}

\newcommand{\mbr}{\mathbold{r}}

\newcommand{\mbu}{\mathbold{u}}
\newcommand{\mbv}{\mathbold{v}}
\newcommand{\mbw}{\mathbold{w}}
\newcommand{\mbx}{\mathbold{x}}
\newcommand{\mby}{\mathbold{y}}
\newcommand{\mbz}{\mathbold{z}}

\newcommand{\mbC}{\mathbold{C}}

\newcommand{\mbT}{\mathbold{T}}
\newcommand{\mbU}{\mathbold{U}}

\newcommand{\mbX}{\mathbold{X}}

\newcommand{\mbZ}{\mathbold{Z}}

\newcommand{\mcC}{\mathcal{C}}
\newcommand{\mcD}{\mathcal{D}}

\newcommand{\mcL}{\mathcal{L}}
\newcommand{\mcM}{\mathcal{M}}

\newcommand{\mcU}{\mathcal{U}}

\newcommand{\mcW}{\mathcal{W}}
\newcommand{\mcX}{\mathcal{X}}
\newcommand{\mcY}{\mathcal{Y}}
\newcommand{\mcZ}{\mathcal{Z}}

\newcommand{\mdE}{\mathds{E}}

\newcommand{\mdP}{\mathds{P}}

\newcommand{\mdR}{\mathds{R}}

\usepackage{tikz}

\makeatletter

\usetikzlibrary{shapes}
\usetikzlibrary{fit}
\usetikzlibrary{chains}
\usetikzlibrary{arrows}

\tikzstyle{latent} = [circle,fill=white,draw=black,inner sep=1pt,
minimum size=20pt, font=\fontsize{10}{10}\selectfont, node distance=1]
\tikzstyle{obs} = [latent,fill=gray!25]
\tikzstyle{const} = [rectangle, inner sep=0pt, node distance=1]
\tikzstyle{factor} = [rectangle, fill=black,minimum size=5pt, inner
sep=0pt, node distance=0.4]
\tikzstyle{det} = [latent, diamond]

\tikzstyle{plate} = [draw, rectangle, rounded corners, fit=#1]
\tikzstyle{wrap} = [inner sep=0pt, fit=#1]
\tikzstyle{gate} = [draw, rectangle, dashed, fit=#1]

\tikzstyle{caption} = [font=\footnotesize, node distance=0] %
\tikzstyle{plate caption} = [caption, node distance=0, inner sep=0pt,
below left=5pt and 0pt of #1.south east] %
\tikzstyle{factor caption} = [caption] %
\tikzstyle{every label} += [caption] %

\newcommand{\plate}[4][]{ %
  \node[wrap=#3] (#2-wrap) {}; %
  \node[plate caption=#2-wrap] (#2-caption) {#4}; %
  \node[plate=(#2-wrap)(#2-caption), #1] (#2) {}; %
}

\makeatother
\usetikzlibrary{backgrounds}
\usetikzlibrary{calc,quotes,arrows.meta}
\usetikzlibrary{shapes.misc,positioning,calc,graphs,arrows}

\pgfdeclarelayer{edgelayer}
\pgfdeclarelayer{nodelayer}
\pgfsetlayers{edgelayer,nodelayer,main}

\definecolor{hexcolor0xbfbfbf}{rgb}{0.749,0.749,0.749}

\tikzset{>=latex}
\tikzstyle{none}   = [inner sep=0pt]
\tikzstyle{line}   = [ -, thick, shorten <=1pt, shorten >=1pt ]
\tikzstyle{arrow}  = [ ->, thick, shorten <=1pt, shorten >=1pt ]
\tikzstyle{ardash} = [ dashed, ->, thick, shorten <=1pt, shorten >=1pt ]

\tikzstyle{box} = [rectangle, minimum width=1.5cm, minimum height=1.5cm,text centered, draw=black, inner sep=7pt]
\tikzstyle{neuron} = [circle, minimum width=4mm, very thick, draw=blue!80!black]

\tikzstyle{empty}=[circle,opacity=0.0,text opacity=1.0,inner sep=0pt]
\tikzstyle{box}=[rectangle,fill=White,draw=Black]
\tikzstyle{filled}=[circle,thick,fill=hexcolor0xbfbfbf,draw=Black]
\tikzstyle{hollow}=[circle,thick,fill=White,draw=Black]
\tikzstyle{param}=[rectangle,fill=Black,draw=Black,inner sep=0pt,minimum width=4pt,minimum height=4pt]
\tikzstyle{paramhollow}=[rectangle,thick,fill=White,draw=Black,inner sep=0pt,minimum
width=4pt,minimum height=4pt]

\usepackage{pgfplots}                               %
\pgfplotsset{compat=newest}
\pgfplotsset{plot coordinates/math parser=false}
\usepgfplotslibrary{statistics}
\newlength\figureheight
\newlength\figurewidth
\newlength\figureheightsmall
\newlength\figurewidthsmall
\definecolor{POSTcolor}{rgb}{0.48, 0.20, 0.58} %
\definecolor{Qcolor}{rgb}{0.00, 0.53, 0.22} %

\makeatletter
\tikzset{
  prefix after node/.style={
    prefix after command={\pgfextra{#1}}
  },
  /semifill/ang/.store in=\semi@ang,
  /semifill/ang=0,
  semifill/.style={
    circle, draw,
    prefix after node={
      \typeout{aaa \semi@ang}
      \let\nodename\tikz@last@fig@name
      \fill[/semifill/.cd, /semifill/.search also={/tikz}, #1]
        let \p1 = (\nodename.north), \p2 = (\nodename.center) in
        let \n1 = {\y1 - \y2} in
        (\nodename.\semi@ang) arc [radius=\n1, start angle=\semi@ang, delta angle=180];
    },
  }
}
\makeatother

\usepackage[normalem]{ulem}
\usepackage{amssymb}

\newcommand\blfootnote[1]{%
  \begingroup
  \renewcommand\thefootnote{}\footnote{#1}%
  \addtocounter{footnote}{-1}%
  \endgroup
}

\colorlet{rct}{orange!90!black}
\definecolor{rwd}{RGB}{31, 119, 180}
\definecolor{miss1}{rgb}{0.0, 0.10980392156862745, 0.4980392156862745}
\definecolor{miss2}{rgb}{0.6941176470588235, 0.25098039215686274, 0.050980392156862744}
\definecolor{miss3}{rgb}{0.07058823529411765, 0.44313725490196076, 0.10980392156862745}

\definecolor{survexternal}{RGB}{17,148,109}
\definecolor{survmatched}{RGB}{207,85,2}
\definecolor{survtrue}{RGB}{67,62,119}
\makeatletter
\tikzset{
    prefix after node/.style={
    prefix after command={\pgfextra{#1}}
  },
  /semifill/ang/.store in=\semi@ang,
  /semifill/ang=0,
  semifill/.style={
    circle, draw,
    prefix after node={
      \typeout{aaa \semi@ang}
      \let\nodename\tikz@last@fig@name
      \fill[/semifill/.cd, /semifill/.search also={/tikz}, #1]
        let \p1 = (\nodename.north), \p2 = (\nodename.center) in
        let \n1 = {\y1 - \y2} in
        (\nodename.\semi@ang) arc [radius=\n1, start angle=\semi@ang, delta angle=180];
    },
  }
}
\makeatother

\begin{document}
\runningauthor{Hau\ss mann, Le, Halla-aho, Kurki, Leinonen, Koskinen, Kaski, L\"ahdesm\"aki}

\twocolumn[

\aistatstitle{Estimating treatment effects from single-arm trials via latent-variable modeling}

\aistatsauthor{Manuel Hau\ss mann${}^{1}$${}$ \And Tran Minh Son Le${}^1$ \And Viivi Halla-aho${}^3$ \And Samu Kurki${}^4$ \AND Jussi V.\ Leinonen${}^4$ \And Miika Koskinen${}^3$ \And Samuel Kaski${}^{1,2}$ \And Harri L\"ahdesm\"aki${}^1$}
\aistatsaddress{\\ ${}^1$Department of Computer Science\\ Aalto University \\  Espoo, Finland \And \\ ${}^2$Department of Computer Science \\ University of Manchester \\   Manchester, United Kingdom \AND ${}^3$ICT Management\\ Helsinki University Hospital\\ Helsinki, Finland \And ${}^4$Bayer Oy \\ Espoo, Finland } 
]

\begin{abstract}
   Randomized controlled trials (RCTs) are the accepted standard for treatment effect estimation but they can be infeasible due to ethical reasons and prohibitive costs.  Single-arm trials, where all patients belong to the treatment group, can be a viable alternative but require access to an external control group. We propose an identifiable deep latent-variable model for this scenario that can also account for missing covariate observations by modeling their structured missingness patterns. Our method uses amortized variational inference to learn both group-specific and identifiable shared latent representations, which can subsequently be used for {\em (i)} patient matching if treatment outcomes are not available for the treatment group, or for {\em (ii)} direct treatment effect estimation assuming outcomes are available for both groups. We evaluate the model on a public benchmark as well as on a data set consisting of a published RCT study and real-world electronic health records. Compared to previous methods, our results show improved performance both for direct treatment effect estimation as well as for effect estimation via patient matching. 
\end{abstract}

\section{INTRODUCTION}\label{sec:intro}

\begin{figure*}
    \centering
        \adjustbox{max width=0.98\textwidth}{
    \begin{tikzpicture}[node distance=1.5cm,>=stealth,plate/.style={draw, shape=rectangle, rounded corners=0.5ex, thick,minimum width=3.1cm, text width=3.1cm, align=right, inner sep=10pt, inner ysep=10pt,append after command={node[above left= 3pt of \tikzlastnode.south east] {#1}}}]
       \node[](covariates){\includegraphics[width=0.22\textwidth, angle=90]{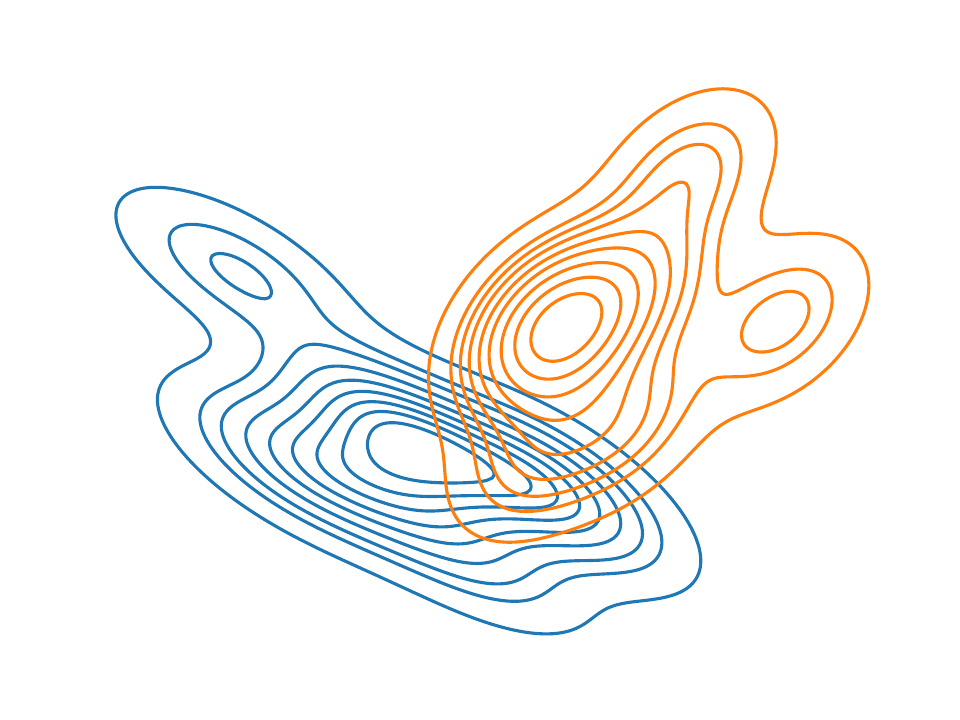}};
       \node[rct,text centered, above=of covariates, yshift=-5.2em, xshift=-2.5em](trial){\emph{single-arm trial data}};
       \node[rwd,text centered, below=of covariates, yshift=5.5em, xshift=-2em](external){\emph{electronic health records}};
       \node[right=of covariates, xshift=7em](latent){\includegraphics[width=0.15\textwidth]{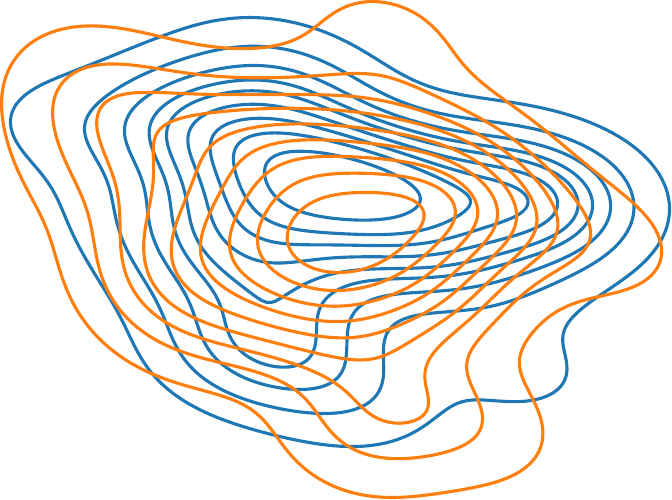}};
       \node[right=of latent, inner sep=0pt, black!80!white, yshift=3.5em, xshift=-10.5em](textshared){\emph{predictive}};

       \node[above=of latent, yshift=-6em,xshift=-9em](poptreat){\includegraphics[width=0.12\textwidth]{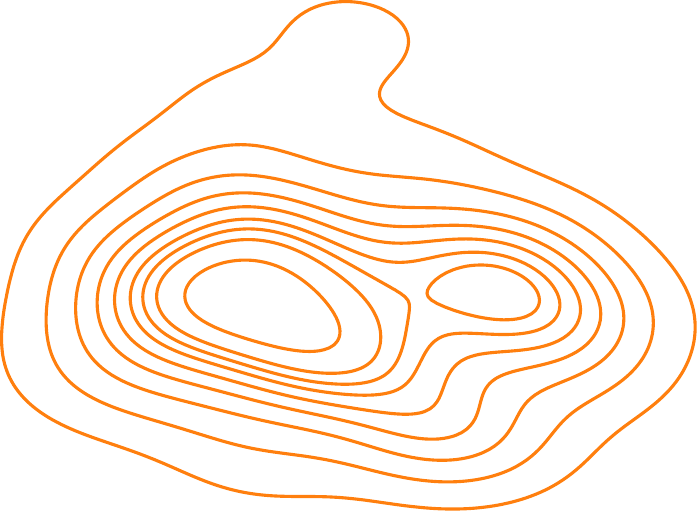}};
       \node[above=of poptreat, inner sep=0pt, rct!90!white, yshift=-4.5em, xshift=1em](texttreat){\emph{treated specific}};
       \node[below=of latent, yshift=6.55em,xshift=-9em](popcontr){\includegraphics[width=0.12\textwidth]{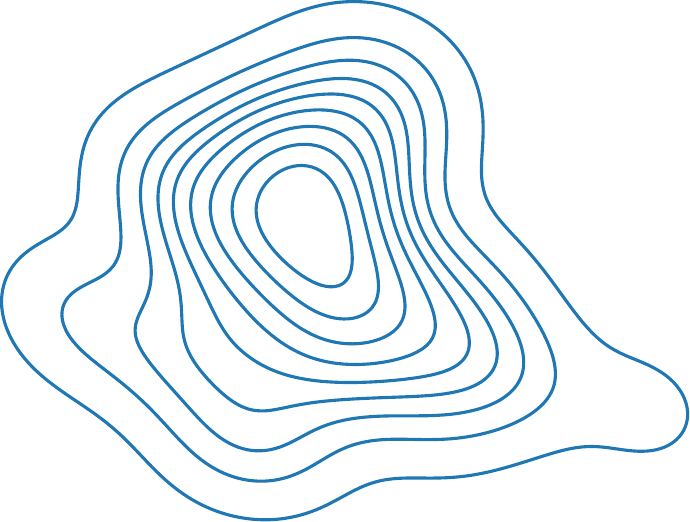}};
       \node[below=of popcontr, inner sep=0pt, rwd!90!white, xshift=0.5em, yshift=4.5em](textcontr){\emph{control specific}};
       \node[rectangle, align=left, right=of latent, ultra thick, rounded corners, fill=green!10!white,draw=green!40!black, inner sep=0.5em, dashed, xshift=-1em] (objective) {\textbf{\color{green!40!black}\small \textsc{Treatment Effect}}\\[0.5em]\quad \E{Y(1) - Y(0)|T=1}};

       \draw[->, ultra thick, rct!60!white,out=0,in=180] (covariates) edge (poptreat);
       \draw[->, ultra thick, rwd!60!white,out=0,in=180] (covariates) edge (popcontr);
       \draw[->, ultra thick, white!60!black,line width=1mm] (covariates) -- (latent);
       \draw[->, ultra thick, white!50!black, in=90] (latent) edge node[above] {\color{white!10!black} \emph{patient matching}} (objective);
       \draw[->, ultra thick, white!60!black, out=-45, in=270, dashed] (latent) edge node[below] {\color{white!60!black}\emph{direct estimate}} (objective);

       \node[plate=\textit{\color{white!10!black} Covariates}, fit=(external) (trial) (covariates), thick, dashed, white!10!black] (covarplate) {};

       \node[plate=\textit{\color{white!10!black} Encodings}, fit=(latent) (popcontr) (poptreat), thick, dashed, white!10!black] (latentplate) {};

\end{tikzpicture}}

    \caption{\emph{Overview.} For the task of {\color{green!40!black}treatment effect estimation} from {\color{rct}single-arm trial data}, patient information for a control group has to be extracted from {\color{rwd}electronic health records} which have been collected, e.g., during hospital visits. Due to their different sources the two covariate distributions only partially overlap. Our model maps them into group-specific latent spaces and a shared, identifiable predictive space. 
    This low-dimensional representation can then be subsequently used for treatment effect estimation. If outcome information is available for both groups, we can obtain a \emph{direct estimate} of the effect from the potential outcomes ($Y(0)$, $Y(1)$).
    We introduce an additional task where treatment outcome information is available only for the control group. In this scenario our method estimates the treatment effect via \emph{patient matching}.
    }
    \label{fig:figureone}
\end{figure*}
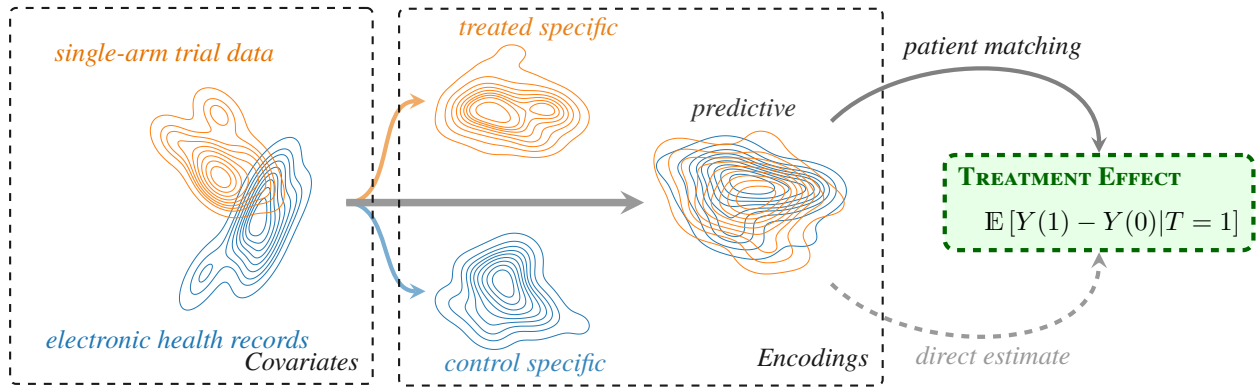

Randomized controlled trials~(RCTs) are the \emph{`gold standard'} in medical research and social sciences for the estimation of treatment effects. 
When conducted properly, RCTs provide control over the treatment assignment and therefore the removal of confounding factors. 
However, this advantage comes with the price of several shortcomings.
Recruiting a sufficient number of patients and collecting their data is usually a costly endeavor both in terms of time as well as financial investments, limiting us to smaller sample sizes in general. 
The study size constraint may further be exacerbated in the case of rare diseases, and RCTs also have ethical challenges in case of serious diseases when no one can be left untreated. 
Single-arm trials offer a way out of these predicaments at the price of requiring external control data, thus no longer being randomized. 

\blfootnote{MH is currently affiliated with \emph{Department of Mathematics and Computer Science; University of Southern Denmark; Odense, Denmark.}}

These control observations are usually taken from the control arms of historical RCTs, or from 
so-called real-world data~(RWD) in the form of electronic health records (EHR) are a promising source for these control observations. 
In the medical domain,
EHR refers to records collected during regular healthcare and hospital visits and not as part of a specific RCT study. 
The degree to which RWD and RCT can replace and complement each other is an open 
research 
question, both theoretical~\citep{collins2020, eichler2020} as well as practical~\citep{franklin2017}. Large-scale projects evaluating trial replicability with RWD data are currently ongoing to better understand their respective strengths and limitations~\citep{dahabreh2020,franklin2021}.

In this work, we consider the task of augmenting single-arm trials with external controls~\citep{gray2020,schmidli2020,chen2021}. 
To estimate treatment effects we have to assume that these two groups have some intrinsic similarity. Yet, given their different sources, their covariate specifics will vary, e.g., concerning the availability of medical history records, breadth of lab measurements, measurement precision, variance in the demographics, etc.
We focus specifically on three problems that arise in this setting. First, due to the differences in the covariate distributions between data collected in RCT and RWD, there is limited overlap between the two sets of observations which has to be overcome to provide a reliable treatment effect estimate~\citep{damour2021}. 
Second, most prior work in the machine learning literature~\citep{johansson2016,shalit2017,shi2019,curth2021flex,bica2022,Jiang2023} has focused on the assumption of having access to treatment outcome information from both the single-arm trial patients as well as the external controls. %
We extend this assumption by considering the additional task of having to find a suitable set of external controls pre-treatment (i.e. before the treatment has been administered to the single-arm group). Here, we have access to pre-treatment covariates for both groups as well as the outcome values for the external control group. 
This setup is highly relevant as it avoids leaking any information from the trial results to the model during the model learning and matching of the patients. Without such new techniques, the inclusion of trial results while inferring the latent variable model could bias further statistical estimates that rely on the learned latent representation.
Finally, we have to account for the fact that real-world data will usually include missing measurements whose non-random patterns have to be accounted for and modeled properly.

We introduce a latent variable model to infer group-specific as well as shared identifiable latent representations for subsequent treatment effect estimation. The former allows us to explain away specifics that are unique to the treatment and the control subsets, respectively, while a shared representation provides a compressed latent space for treatment effect estimation as well as patient matching, i.e., the selection of a subset of control patients that are most similar to the treated group.
This differs from the popular approach in the medical literature that relies on various forms of propensity score estimators~\citep{stuart2010}, which do not infer such a space prior to estimating a matching score.
The problem setting and our approach are conceptualized in~\cref{fig:figureone}.

\paragraph{Contributions.} 
In this work, we consider the task of estimating treatment effects from single-arm trials with external controls and also introduce an additional scenario where outcome information is not available for the treatment group during inference. We contribute
\begin{itemize}
    \item[\emph{(i)}] a principled way of handling these tasks via amortized latent-variable models with identifiability guarantees that can infer a predictive latent space between two different covariate distributions for subsequent treatment effect prediction, and simultaneously model structured missingness patterns;
    \item[\emph{(ii)}] an extensive ablation study demonstrating that our method is competitive if outcome information is available for both treated and control groups, as well as if it is available only for the latter. The method improves upon prior work on several variations of a semi-synthetic benchmark as well as on a curated large-scale data set that combines data from a published RCT study \citep{Bakris2020} with real-world EHRs.
\end{itemize}

\section{RELATED WORK}\label{sec:related}

\paragraph{Treatment Effect Estimation.}
Treatment effect estimation from observational studies~(OS) has a long history 
See, e.g., \citet{imbens2004} for a review on \emph{average treatment effect (ATE)} estimation approaches used within the statistics and econometrics literature, which rely on a variety of machine learning approaches, e.g., random forests~\citep{hill2011,athey2019}.
Given the increasing amount of observational data available, 
a recent trend in the machine learning literature has been to focus on models for \emph{individual treatment effect} or \emph{conditional average treatment effect~(CATE)} estimations~(see, e.g., \cite{bica2021} for an introduction). 
A broad spectrum of  approaches exists to construct deterministic CATE estimators \citep{Kunzel2019}.
See \citet{curth2021} for a recent overview and unifying framework to classify different neural network-based approaches.
A common approach is to build upon the theory of representation learning~\citep{bengio2013} to construct deterministic representations that are then used by task-specific mappings~\citep{johansson2016,shalit2017,shi2019,Jiang2023}. 
In parallel, there is an increasing interest in using generative models for the task of CATE estimation, relying, e.g., on variational autoencoders~\citep{kingma2014,louizos2017,lu2020,Zhang2021}, generative adversarial networks~\citep{goodfellow2014,yoon2018}, or energy-based approaches~\citep{lecun2006,zhang2022}.
Our method belongs to this second group of generative approaches, focuses on amortized variational inference, and extends prior work to the specific task of learning with external controls in settings that can have strongly divergent groups of samples.

\paragraph{Combining RCT with OS.} 
Combining small randomized control trial data sets with observational data has been the focus of several studies in recent years. 
It can be seen as a form of domain adaptation ~\citep{ben2010} as the aim is to combine different data sources with similar yet distinct characteristics. 
\citet{kallus2018} consider combining the two to get rid of hidden confounders requiring the very restrictive assumption of linear biases. 
\citet{cheng2021} propose to learn separate estimators,
combining them in a second step via a weighting scheme. However, finding these weights requires a separate validation set. 
\citet{hatt2022} propose a two-step approach in which they first rely on learning a preliminary representation solely from observational study data, which they aim to fine-tune by learning data-specific structures via the randomized trial data. 
While these approaches focus on the combination of two complete studies, our setting focuses on single-arm information from each of our groups. 
Closest to our formulation are \citet{bica2022}, who consider combining multiple heterogeneous data sources into a joint model for CATE estimation, requiring full treatment outcome information for all data sources. 

\paragraph{External Controls and Patient Matching.} 
Matching consists of pairing each patient in the treatment group with a suitable patient from the set of external controls, based on a similarity score, and has long been a topic of ongoing research with a wide spectrum of approaches~\citep{stuart2010}. 
E.g., \citet{li2017} rely on kernel methods and maximum mean discrepancy regularization to construct latent spaces for subsequent matching. \citet{luo2020}, in turn, rely on central subspaces. \citet{athey2018} can approximately ensure balancing between the covariates in high-dimensional spaces under the assumption of only linear biases between them.

\paragraph{Unconfoundedness and Identifiability.}
A common assumption for causal inference is the absence of hidden confounders. To avoid this assumption prior works rely, e.g., on having multiple causes \citep{wang2019}, access to multiple treatments over time \citep{bica2020}, or by relying on invariant risk minimzation \citep{arjovsky2019, shi2019}.
Obtaining guarantees for identifiability of learned latent spaces is similarly an area of ongoing research \citep{Xi2023, moran2022}. Our approach relies on using auxiliary variables following the work by \citet{khemakhem2020}.

\paragraph{Estimation under Missingness.}
Modeling with missing treatment outcome information has been pursued primarily with the goal of average treatment effect estimation~\citep{williamson2012,zhang2016,kennedy2020}. \citet{kuzmanovic2022} generalize such prior approaches by focusing on building CATE estimators with missing treatment information. 
We differ from these as we do not assume randomly missing outcome data but are restricted to \emph{systematically} missing treatment outcome information from the treatment group. This is necessary when we want to match a set of external control patients with a group of patients recruited specifically for the study at hand.
Additionally, in real-world applications, we are often faced with missing covariate information~\citep{PerezLebel2022}. 
Modeling missing not at random (MNAR) covariates~\citep{Rubin1976} has received a lot of focus in the generative modeling literature recently~\citep{collier2020vaes,ipsen2021,ghalebikesabi21a}. 
Throughout this work we build on the work by \citet{collier2020vaes}.

\section{BACKGROUND}\label{sec:background}

\paragraph{Problem Specification.} 
We have access to a sample of $N$ patients with covariates $\mbx \in \mcX$, outcomes~${y \in \mcY}$ and treatment assignments $t \in \{0,1\}$, such that ${\mcD = \{(\mbx_1,y_1, t_1),\ldots,(\mbx_N,y_N,t_N)\}}$. 
Following \citet{rubin2005}'s potential outcomes formulation we assume potential outcomes $Y_n(0), Y_n(1) \in \mcY$ of which due to the fundamental problem of causality~\citep{pearl2009} only one can be observed, such that 
$y_n = t_nY_n(1) + (1 - t_n) Y_n(0)$.

Throughout, we have to assume three standard assumptions necessary for estimating causal effects from observational data: 
\emph{(i)~consistency}, i.e., if patient $n$ has received treatment $t_n$ we observe potential outcome ${y_n = Y_n(t_n)}$; 
\emph{(ii)~unconfoundedness}, $Y(0),Y(1)\indep T | X$, i.e., there are no unobserved confounders; 
\emph{(iii)~overlap}, ${0<\pi(x)<1\;\forall x}$, i.e., the treatment assignment is not deterministic, where 
    $\pi(x) \deq \mdP(T=1|X=x)$,
is known as the \emph{propensity score}. 
Assumptions {\em (ii)} and {\em (iii)} provide us with tension as they can be difficult to simultaneously fulfill~\citep{damour2021}. To ensure unconfoundedness, including an increasing number of covariates is usually considered to be helpful. However, to ensure overlap, a low-dimensional space is desired. We see below how this tension is handled within our model.
Given the diverse nature of the treatment and control groups in our setup, this third constraint is the most critical and requires careful consideration.

Observing outcomes of both arms, we estimate the \emph{conditional average treatment effect (CATE)},
\begin{equation}\label{eq:CATE}
    \tau_{C}(x) \deq \E{Y(1) - Y(0)|X=x} = \mu_1(x) - \mu_0(x),
\end{equation}
where $\mu_t(x) \deq \E{Y(t)|X=x}$. 
The \emph{average treatment effect (ATE)} is then defined as $\E{\tau_C(X)}$.

The main focus of our work is the case where outcomes are only available for the control group, in which case 
we consider the \emph{average treatment effect for the treated~(ATT)}, 
\begin{equation}\label{eq:att}
    \tau_{A}\deq \E{Y(1) - Y(0)|T=1}, 
\end{equation}
and estimate it with a suitable control group by matching external control patients with the treated.

\section{OUR MODEL}\label{sec:model}

We introduce a latent-variable model for the task of treatment effect estimation from single-arm trials with external controls focusing also on the task where outcome information is only available for the control group.

\paragraph{A Generative Model.}
Previous models \citep{louizos2017,lu2020} have so far solely focused on learning latent embeddings within the setting of having observed outcome information for both the treatment and the control groups. From their joint
\begin{equation}
p(t,\mbx,\mby,\mbz) = p(t|\mbz)p(\mbx|\mbz)p(y|t,\mbz)p(\mbz),\label{eq:basicjoint}
\end{equation}
we notice the dual role played by the latent representation~$\mbz$. %
In~\eqref{eq:basicjoint} $\mbz$ is assumed to be predictive of $y$ and, at the same time, to offer a reliable encoding of~$\mbx$. These tasks counteract each other. 
Accurately modeling  $p(y|t,\mbz)$ requires extracting predictive information from the covariates $\mbx$ (as~$\mbz$ is latent), which we assume to sufficiently overlap between the treatment and the control groups. Modeling $p(\mbx|\mbz)$, however, forces the model to encode a complete representation, i.e., to take covariate information into consideration that is irrelevant to the predictive task, and to further model group-specific variations existing in the data. This was observed to be detrimental, e.g., by \citet{lu2020}, who dropped it from their final objective. 

\begin{figure}
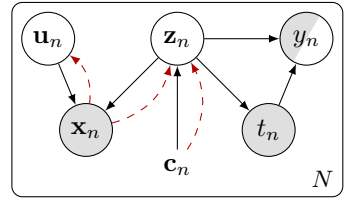

   \centering
   \tikz{
       \colorlet{infer}{red!70!black}
        \node[latent] (z) {$\mbz_n$};
        \node[obs, below left=of z] (x) {$\mbx_n$};
        \node[semifill={gray!25, ang=60}, draw=black,inner sep=1pt, minimum size=20pt, font=\fontsize{10}{10}\selectfont, node distance=1, right=of z] (y) {$y_n$};
        \node[latent, left=of z] (u) {$\mbu_{n}$};
        \node[obs, below right=of z] (t) {$t_n$};
        \node[below of=z, yshift=-2em] (c) {$\mbc_n$};
        \draw[->] (z) to (x);
        \draw[->] (z) to (t);
        \draw[->] (z) to (y);
        \draw[->] (u) to (x);
        \draw[->] (t) to (y);
        \draw[->] (c) to (z);
        \draw[->, dashed, infer] (c) to[bend right] (z);
        \draw[->, dashed, infer] (x) to[bend right] (z);
        \draw[->, dashed,infer] (x) to[bend right] (u);
        \plate{}{(u)(x)(y)(t)(z)}{$N$};
   }
   \caption{\emph{Plate Diagram of Our Model.} Black solid arrows denote the generative model, red dashed arrows the inferential dependency. Empty, partially filled, and filled circles refer to latent, partially, and fully observed variables. 
   }\label{fig:plate}
\end{figure}
We take these different tasks into account and reduce the tension by extending the original latent space representation by adding an additional latent variable $\mbu \in \mcU$ to the joint distribution.
This variable aims to explain away group-specific characteristics, allowing $\mbz$ to be predictive of the treatment, yet still be guided by the reconstructive task it performs jointly with $\mbu$ through the resulting likelihood $p(\mbx|\mbu,\mbz)$.
However, as deep unconstrained latent-variable models are not identifiable \citep{khemakhem2020} $\mcZ$ is also not identifiable due to $p(\mbz)$ being unconditional. We instead have to rely on a conditional prior $p(\mbz|\mbc)$, where~$\mbc$ is an additional observed variable, separate from the covariates $\mbx$. In practice, $\mbc$ contains covariates that are best considered as non-random variables, such as gender, age, country of origin, etc. 
With a normal prior for $\mbu$, our model (see \cref{fig:plate}) is given as
\begin{align*}
    \mbu &\sim p(\mbu) = \Norm(\mbu|0, \sigma_u^2\1),\\
    \mbz &\sim p(\mbz|\mbc) = \Norm(\mbz|\lambda(\mbc), \sigma_z^2\1),\\
    \mbx &\sim p(\mbx|\mbu, \mbz),\\
    t &\sim p(t|\mbz) = \Ber\big(t|\sigma(f(\mbz))\big),\\
    y &\sim p(y|t, \mbz)= \Norm\big(y|\mu^y_t(\mbz), \alpha^{-1}\big),
\end{align*}
where $\lambda(\cdot), f(\cdot)$ and $\mu_t^y(\cdot)$ are neural nets, $\alpha$ is a precision parameter, ${\sigma(x) \deq 1/(1 + \exp(-x))}$ the logistic sigmoid, and $\1$ an identity matrix. %
We assume the likelihood $p(\mbx|\mbu,\mbz)$ to factorize over the covariates, i.e.,
${p(\mbx|\mbu,\mbz) = \prod_{i=1}^{d_x}p(x_i|\mbu, \mbz)}$,
where ${d_x \deq \dim(\mbx)}$. In the experiments, we model the individual covariates $x_i$ via Bernoulli, Categorical, or Normal distributions, depending on their domain. For continuous covariates, we assume a homoscedastic noise model, i.e., $\Norm(x_i|g(\mbz)_i,\beta^{-1}_i)$ with precision parameters~$\beta_i$. The precisions $\alpha$, and $\{\beta_i\}_i$ are optimized together with the remaining model parameters. 
Assuming $\mcU$ is high-dimensional enough, we can rely on it to model the group-specific latent spaces of both the treatment and control groups. 
Alternatively, they can be explicitly separated into group-specific latent spaces, such that $\mcU = [\mcU_0, \mcU_1]$. This allows for an easy extension of the current two-group setup to multiple groups which may only become available at a later stage. We explore this variant in our experiments but do not find the additional separation to be necessary in most cases.

\paragraph{Inference.} 
We rely on amortized variational inference \citep{kingma2014, rezende2014}
to infer an approximation to the intractable posterior $p(\mbU,\mbZ|\mcD)$, where ${\mbU = (\mbu_{1},\ldots, \mbu_{N})}$, 
 $\mbC$, $\mbX$, $\mbZ$ analogously.
We assume a mean-field variational posterior
\begin{align*}
q(\mbU,\mbZ|\mbC,\mbX) &= \prod_{n=1}^N q(\mbu_{n}|\mbx_n)q(\mbz_n|\mbc_n,\mbx_n ), \quad\text{where}\\
q(\mbu_n|\mbx_n) &= \Norm\big(\mbu_{n}|\mu^u(\mbx_n), \sigma^u(\mbx_n)\big) \text{ and}\\
q(\mbz_n|\mbc_n,\mbx_n) &= \Norm\big(\mbz_n|\mu^z(\mbc_n,\mbx_n), \sigma^z(\mbc_n,\mbx_n)\big),
\end{align*}
with neural networks $\mu^u(\cdot)$, $\mu^z(\cdot)$, $\sigma^u(\cdot)$, and $\sigma^z(\cdot)$. 
We minimize the Kullback-Leibler divergence between the two, i.e., $\KL{q(\mbU,\mbZ|\mbC,\mbX)}{p(\mbU,\mbZ|\mcD)}$,
by maximizing the corresponding \emph{evidence lower bound (ELBO)} with respect to all parameters. If outcome values are available for both  groups, the ELBO is given as
\begin{align}\label{eq:elbo}
    \log~&p(\mcD) \geq \sum_{n=1}^N \Ep{q(\mbu_{n},\mbz_n|\mbc_n, \mbx_n)}{\log p(\mbx_n|\mbu_{n},\mbz_n)}\nonumber \\
    &\quad+ \mdE_{q(\mbz_n|\mbc_n, \mbx_n)}\Big[\underbrace{\log p(y_n|t_n,\mbz_n)}_{(\star)} + \log p(t_n|\mbz_n)\Big]\nonumber\\
                 &\quad -\KL{q(\mbu_{n}|\mbx_n)\big.}{p(\mbu_{n})}\nonumber \\
                 &\quad- \KL{q(\mbz_n | \mbc_n,\mbx_n)\big.}{p(\mbz_n|\mbc_n)} \deq \mcL_\text{elbo}^\text{full}.
\end{align}
If outcome values $y_n$ are available only for the control group, we optimize a modified ELBO.
The missing outcome for the single-arm treatment group is masked out, i.e., we replace $(\star)$ in the equation above with 
\begin{equation*}
\Ep{q(\mbz_n|\mbc_n,\mbx_n)}{(1 - t_n)\log p(y_n|t_n,\mbz_n)}.    
\end{equation*}
See \cref{sec:app-theory} for further details. 

We follow prior work \citep{johansson2016,shalit2017,lu2020} and rely on tools from domain adaptation \citep{ben2010} to further constrain $\mbz$. Specifically, we rely on the gradient-reversal layer approach by \citet{ganin2015} providing us with a generative adversarial network-based regularizer \citep{goodfellow2014}. 
Our model tries to fool a discriminator whose task is to distinguish between the encoded samples from the two groups, i.e., whose aim is to maximize $\log p(t_n|\mbz_n)$. 
See \cref{sec:app-expdetails} for details on the modified objective.

\paragraph{Identifiability.}
As mentioned above, fully unconstrained latent-variable models, ${p_\theta(\mbx,\mbz) = p_\theta(\mbx|\mbz)p_\theta(\mbz)}$, are non-identifiable in the sense that the implication 
\begin{equation*}
    \forall (\theta, \theta'): \quad p_\theta(\mbx) = p_{\theta'}(\mbx) \Rightarrow \theta =\theta',
\end{equation*}
does not necessarily hold \citep{khemakhem2020}, i.e., different model parameterizations can lead to the same marginal. 
\citet{khemakhem2020} propose to constrain the model via an additional set of variables $\mbc$ and assume a joint distribution that factorizes as 
\begin{equation*}
    p_\theta(\mbx,\mbz|\mbc) = p_\theta(\mbx|\mbz)p_\theta(\mbz|\mbc),
\end{equation*}
where $\mbx = f_\theta(\mbz) + \veps$, with $\veps\sim p(\veps)$ being an independent noise variable. 
The prior $p(\mbz|\mbc)$ is assumed to be parameterizable as a factorizing exponential family,
\begin{equation}\label{eq:condprior}
    p(\mbz|\mbc) = \prod_{i=1}^{d_z}m(z_i)/Z(\mbc)\exp\left(T(z_i)^\top\lambda(\mbc)\right).
\end{equation}
Assuming a sufficiently diverse set of variables $\mbc \in \mcC$, there need to be $d_zk+1$ distinct values, where $k$ is the number of sufficient statistics, then $\theta \deq (f,T,\lambda)$ is identifiable up to permutations and translations. 
\citet{Xi2023} later showed that the assumption on a factorizing exponential family is not necessary and that, due to the requirement on the independent noise variable $\veps$ in $p(\mbx|\mbz)$ theoretical guarantees only hold for continuous covariates~$\mbx$.
Empirical results by \citet{khemakhem2020} indicate that identifiability still tends to be achievable for discrete covariates. 
Throughout our experiments, whenever we encounter a mixture of discrete and continuous covariates, we still rely on a prior as specified in \eqref{eq:condprior} and observe improved results, while theoretical guarantees could be maintained by ignoring discrete covariates within the likelihood.

Within our setup, we are interested in identifying the predictive latent space of $\mbz$. We keep the prior $p(\mbu)$ unconstrained and model $\mbz$'s conditional prior $p(\mbz|\mbc)$ as a factorized normal distribution, whose $\lambda(\cdot)$ are parameterized by a neural net. 
The argument by \citet{khemakhem2020} then guarantees identifiability of $\mbz$ up to the constraints mentioned above. 
See \cref{sec:app-ident} for a detailed discussion on the required conditions and derivations.

\paragraph{Patient Matching.}
Matching consists of extracting a subset of observations from the control group that is similar to the observations in the single-arm treatment group with respect to a feature~$s$. 
Such a feature is often created via the propensity score.\footnote{A classical result by \citet{rosenbaum1983} proves that if the average treatment effect is identifiable from observational data after adjusting for $X$, then adjusting for the propensity score $\pi(X)$ is sufficient.} Given $s$, a distance measure
estimates the similarity between two observations. That is, the matching control $\mbx_\text{match}^i$ for an observation $\mbx_i$ with $s_i$ is 
\begin{equation*}
    \mbx_\text{match}^i = \mbx_j, \quad\text{with } j = \argmin_{j \in J_c} d(s_i, s_j),
\end{equation*}
where $d(\cdot,\cdot)$ is a distance metric and $J_c=\{j|t_j=0\}$ is the index set over the control observations. 

As $\mcZ$ is designed to extract a predictive encoding, we perform patient matching within this compressed, low-dimensional latent space,
and match in $\mcX$ only for ablation purposes.
If we consider the inferred posterior means in $\mcZ$ as point estimates that encode the covariates, we  either $\emph{(i)}$ estimate a propensity score given the encodings $\mu^z(\mbx_i)$, i.e., ${s_i = \hat\pi(T=1|\mu^z(\mbx_i))}$, \emph{(ii)} use the posterior mean directly, i.e., $s_i = \mu^z(\mbx_i)$, or \emph{(iii)} the full variational posterior, i.e., $s_i = (\mu^z(\mbx_i), \sigma^z(\mbx_i))$.
Given similarity distances, we rely on nearest-neighbor matching with replacement. This straightforward approach could be extended to further improve the overall overlap by taking additional measures into account, such as, e.g., calipers~\citep{austin2011}. Throughout the experiments, we use the Euclidean distance for $d(\cdot,\cdot)$. 
See \cref{sec:app-metric} for matching without replacement and an ablation using distributional distance measures.

\paragraph{Missingness.}
So far, our model assumes fully observed covariates $\mbx$ --- an assumption that is often violated in practice. We generally cannot assume that EHRs have measurements of all the features we require in a specific study for every patient. 
Additionally, we cannot assume that missing measurements are completely random but rather have to assume dependency structures within the missingness, i.e., missing not at random (MNAR)~\citep{Rubin1976}.\footnote{While MNAR is our focus in this work, the generative model could easily be adapted to missing (completely) at random assumptions as well.}
To account for this we build upon the proposal by \citet{collier2020vaes}. 
Given masking variables ${\mbm \in \mcM = \{0,1\}^{d_x}}$, where $\mbm_{ni}=1$ indicates that the $i$-th covariate has been observed in the $n$-th sample, we observe masked  covariates ${\tilde \mbx = \mbm \circ \mbx + (1 - \mbm) \circ \boldsymbol{\eta}}$, for some imputation $\boldsymbol{\eta}$ 
and~$\circ$~the element-wise product. 
To account for the MNAR structure, we model $\mbm$ to depend on an additional latent variable~$\mbz^m \sim p(\mbz^m)$ as well as $(\mbu, \mbz)$ such that
\begin{align*}%
    \mbm|\mbz^m, \mbu, \mbz &\sim \prod_{i=1}^{d_x}\Ber\big(m_i|\sigma(g(\mbz^m, \mbu, \mbz)_i)\big), \\
    \mbx|\mbm, \mbu,\mbz^x &\sim p(\mbx|\mbm, \mbu,\mbz^x).\nonumber
\end{align*}
See \cref{sec:app-missing} for the complete model.

\section{EVALUATION}\label{sec:evaluation}

\begin{table*}
    \centering
    \caption{\emph{Full \& Partial Outcome Observation.} Variations of modified IHDP covariates and simulated treatment outcomes.
    Our proposal improves on both deterministic and probabilistic baselines. 
    (`--' refers to a method not being applicable)
    }\label{tab:main-joint}
        \adjustbox{max width=0.98\textwidth}{
    \begin{tabular}{lcccccccccc}
    \toprule
    &\multicolumn{4}{c}{(a) full outcome observation (RMSE of CATE)} & \multicolumn{4}{c}{(b) partial outcome observation (AE of ATT)}\\ 
    \cmidrule(lr){2-5} \cmidrule(lr){6-9}
    &\multicolumn{2}{c}{all+high}  & \multicolumn{2}{c}{subset+low}&\multicolumn{2}{c}{all+high}  & \multicolumn{2}{c}{subset+low}  \\
    \cmidrule(lr){2-3}\cmidrule(lr){4-5}\cmidrule(lr){6-7}\cmidrule(lr){8-9}
\textsc{Method} & within sample & out-of-sample & within sample& out-of-sample& within sample & out-of-sample & within sample& out-of-sample \\
    \midrule
    CFor & $0.556 \sd{\pm 0.000}$ & $0.5967 \sd{\pm 0.001}$  & $3.449\sd{\pm 0.052}$ & $2.597 \sd{\pm 0.031}$&--&--&--&--\\
    PScov &--&--&--&--& $0.115 \sd{\pm0.005}$ &$0.336 \sd{\pm0.017}$  &$0.691 \sd{\pm0.059}$ &$0.702 \sd{\pm0.063}$\\
    PSpca &--&--&--&--& $0.096 \sd{\pm0.005}$ &$0.300 \sd{\pm0.014}$ &$0.496 \sd{\pm0.036}$ &$0.613 \sd{\pm0.051}$\\
    PSlat &--&--&--&--& $0.084 \sd{\pm0.004}$ &$0.187 \sd{\pm0.013}$  &$0.276 \sd{\pm0.025}$ &$0.272 \sd{\pm0.023}$\\
    SingleNet & $0.241 \sd{\pm 0.005}$ &$0.328 \sd{\pm 0.009}$  &$0.924 \sd{\pm 0.016}$ &$0.878 \sd{\pm 0.027}$&--&--&--&--\\
    TNet & $0.175 \sd{\pm 0.003}$ &$0.275 \sd{\pm 0.008}$  &$0.278 \sd{\pm 0.009}$ &$0.328 \sd{\pm 0.017}$&--&--&--&--\\ 
    TARNet    & $0.177 \sd{\pm 0.004}$ &$0.280 \sd{\pm 0.008}$  &$0.272 \sd{\pm 0.009}$ &$0.333 \sd{\pm 0.019}$ & $0.043 \sd{\pm0.003}$ &$0.130 \sd{\pm0.008}$  &$0.140 \sd{\pm0.011}$ &$0.239 \sd{\pm0.016}$ \\
    CFRNet  & $0.171 \sd{\pm 0.003}$ &$0.279 \sd{\pm 0.008}$  &$0.279 \sd{\pm 0.008}$ &$0.338 \sd{\pm 0.017}$& $0.042 \sd{\pm0.002}$ &$0.131 \sd{\pm0.007}$ &$0.345 \sd{\pm0.016}$ &$0.370 \sd{\pm0.017}$ \\
    SNet & $0.168 \sd{\pm 0.003}$ &$0.264 \sd{\pm 0.008}$  &$0.211 \sd{\pm 0.008}$ &$\underline{\mathbf{0.243 \sd{\pm 0.016}}}$& $0.043 \sd{\pm0.003}$ &$0.137 \sd{\pm0.007}$  &$0.221 \sd{\pm0.015}$ & $0.366 \sd{\pm0.026}$\\
    VAE &--&--&--&--& $0.085 \sd{\pm0.004}$ &$0.266 \sd{\pm0.012}$  &$0.264 \sd{\pm0.014}$ &$0.427 \sd{\pm0.022}$ \\
    CEVAE  &$0.182 \sd{\pm 0.002}$ &$0.287 \sd{\pm 0.007}$ &$0.288 \sd{\pm 0.009}$ &$0.350 \sd{\pm 0.017}$& $0.052 \sd{\pm0.002}$ &$0.174 \sd{\pm0.009}$  &$0.150 \sd{\pm0.009}$ &$0.259 \sd{\pm0.015}$ \\
    TEDVAE & $0.176 \sd{\pm 0.002}$ &$0.283 \sd{\pm 0.007}$ &$0.293 \sd{\pm 0.009}$ &$0.336 \sd{\pm 0.018}$ &$0.052 \sd{\pm0.003}$ &$0.179 \sd{\pm0.009}$ &$0.144 \sd{\pm0.009}$ & $0.272 \sd{\pm0.016}$ &\\
    \midrule
    Ours   & $0.152 \sd{\pm 0.002}$ &$0.267 \sd{\pm 0.008}$  &$0.201 \sd{\pm 0.007}$ &$0.272 \sd{\pm 0.016}$& $\mathbf{0.037 \sd{\pm0.002}}$ & $\mathbf{0.113 \sd{\pm0.007}}$  &$\mathbf{0.114 \sd{\pm0.008}}$ &$\mathbf{0.190 \sd{\pm0.016}}$ \\
    \quad+I   &$0.141 \sd{\pm 0.002}$ &${0.251 \sd{\pm 0.008}}$  &$0.194 \sd{\pm 0.007}$ &$0.262 \sd{\pm 0.016}$& $\mathbf{0.038 \sd{\pm0.002}}$ &$\mathbf{0.122 \sd{\pm0.007}}$  &$\mathbf{0.110 \sd{\pm0.007}}$ &$\underline{\mathbf{0.176 \sd{\pm0.014}}}$\\
    \quad+sep &$0.151 \sd{\pm 0.002}$ &$0.267 \sd{\pm 0.008}$  &$0.209 \sd{\pm 0.007}$ &$0.278 \sd{\pm 0.016}$& $\mathbf{0.037 \sd{\pm0.002}}$ &$\mathbf{0.111 \sd{\pm0.007}}$ &${0.117 \sd{\pm0.009}}$ &$\mathbf{0.193 \sd{\pm0.016}}$\\
    \quad+sep+I &$0.143 \sd{\pm 0.002}$ &$0.255 \sd{\pm 0.007}$ &$0.197 \sd{\pm 0.007}$ &$0.266 \sd{\pm 0.017}$ & $\underline{\mathbf{0.035 \sd{\pm0.002}}}$ &$\mathbf{0.117 \sd{\pm0.008}}$ &${\mathbf{0.104 \sd{\pm0.007}}}$ &${\mathbf{0.185 \sd{\pm0.015}}}$\\
    \quad+snet & $0.141 \sd{\pm 0.002}$ &$0.256 \sd{\pm 0.008}$ &$\mathbf{0.185 \sd{\pm 0.007}}$ &$\mathbf{0.253 \sd{\pm 0.017}}$ & $\mathbf{0.038 \sd{\pm0.002}}$ &$\mathbf{0.117 \sd{\pm0.007}}$ &$0.194 \sd{\pm0.014}$ &$0.322 \sd{\pm0.020}$ &\\
    \quad+snet+I & $\underline{\mathbf{0.130 \sd{\pm 0.002}}}$ &$\underline{\mathbf{0.242 \sd{\pm 0.008}}}$ &$\mathbf{0.181 \sd{\pm 0.006}}$ &$\mathbf{0.250 \sd{\pm 0.017}}$ &$\mathbf{0.038 \sd{\pm0.002}}$ &$\mathbf{0.120 \sd{\pm0.008}}$ &$0.190 \sd{\pm0.012}$ &$0.322 \sd{\pm0.023}$ &\\
    \quad+snet+sep & $0.143 \sd{\pm 0.002}$ &$0.254 \sd{\pm 0.008}$  &$0.191 \sd{\pm 0.007}$ &$0.258 \sd{\pm 0.017}$& $0.040 \sd{\pm0.003}$ &$\underline{\mathbf{0.109 \sd{\pm0.007}}}$  &$0.179 \sd{\pm0.013}$ &$0.316 \sd{\pm0.020}$ \\
    \quad+snet+sep+I & ${\mathbf{0.131 \sd{\pm 0.002}}}$ &${\mathbf{0.247 \sd{\pm 0.008}}}$ &$\underline{\mathbf{0.180 \sd{\pm 0.007}}}$ &${\mathbf{0.251 \sd{\pm 0.017}}}$& $0.039 \sd{\pm0.003}$ &$\mathbf{0.115 \sd{\pm0.007}}$  & $0.196 \sd{\pm0.015}$ &$0.292 \sd{\pm0.020}$\\
    \quad+tedvae & $0.143 \sd{\pm 0.002}$ &$0.257 \sd{\pm 0.008}$ &$\mathbf{0.187 \sd{\pm 0.007}}$ &$0.258 \sd{\pm 0.019}$ &$\mathbf{0.039 \sd{\pm0.003}}$ &$\mathbf{0.119 \sd{\pm0.007}}$ &$\mathbf{0.103 \sd{\pm0.008}}$ &$\mathbf{0.189 \sd{\pm0.015}}$ &\\
    \quad+tedvae+I & $0.142 \sd{\pm 0.002}$ &$0.255 \sd{\pm 0.008}$ &$0.191 \sd{\pm 0.006}$ &$\mathbf{0.253 \sd{\pm 0.016}}$ &$\mathbf{0.036 \sd{\pm0.002}}$ & $0.124 \sd{\pm0.007}$ & $\underline{\mathbf{0.102 \sd{\pm0.007}}}$ &$\mathbf{0.191 \sd{\pm0.017}}$ &\\
    \bottomrule
    \end{tabular}}\\
    \tiny{\emph{mean $\pm$ standard error over 300 random replications; statistically significant best models marked \textbf{bold}}; lowest mean \underline{underlined}}
\end{table*}

\paragraph{Data Sets.}
Given the fundamental problem of causality, we have to rely on semi-synthetic data sets to evaluate the efficacy of the proposed method. 
We first consider adaptations of the commonly used semi-synthetic IHDP benchmark~\citep{hill2011} consisting of 25 real covariates (one categorical, 19 binary, five continuous) and synthetic continuous treatment outcomes $y$.
Additionally, we create a survival analysis setup where $y$ represents time-to-event information, with the aim of estimating hazard ratios.
As a second real-world data set, we combine $140$,  partially missing, covariates (100 binary, 40 continuous) of $833$ participants of a published RCT study \citep{Bakris2020} with real-world electronic health record~(EHR) data from $2646$ patients. 
Details on these data sets as well as further experimental details not mentioned in the subsequent subsections are discussed in~\cref{sec:app-expdetails}. 
As far as possible, all variants share the same architectures and hyperparameters. 
See~\cref{sec:app-eval} for further evaluations. 
An implementation of our proposed approach is available at \url{https://github.com/manuelhaussmann/lvm_singlearm}.

\paragraph{Experimental Assumptions.} 
We consider two cases. Either \emph{(a) outcome information is available for both groups}, i.e., the single-arm trial group and the external control group, which is the common assumption used in the deep learning literature. In this case, estimators can be inferred directly. Or \emph{(b) outcome information is only available for the control group}. Here, inference of a representation space is required, which can be used in a second step to select a suitable subset of control patients via matching.
We evaluate scenarios where either \emph{all} covariates are potentially predictive, or only a \emph{subset} of them, and where the covariate overlap between the two groups is either \emph{high}, where we keep the original covariates, which are already only partially overlapping, or \emph{low}, where we additionally shift them to decrease their overlap even further.

\paragraph{Baselines \& Ablations.}
We compare our method to several baselines. 
\emph{SingleNet} serves as our simplest deterministic baseline and learns a single estimator $\mu(x,t)$. 
\emph{TNet} generalizes this to two separate neural net-based estimators $\mu_t(x)$. 
\emph{TarNet} and \emph{CFRNet} in turn learn a shared representation space from which they predict treatment outcomes \citep{shalit2017}. 
\emph{SNet} further generalizes this by learning a mixture of unique and shared representations \citep{curth2021}.
\emph{VAE} serves as our vanilla generative baseline following the standard architecture proposed by \citet{shalit2017}. 
\emph{CEVAE} extends it with an additional treatment outcome estimator \citep{louizos2017}. 
Finally, \emph{TEDVAE} generalizes CEVAE by splitting the latent space into separate predictive parts similar to the deterministic SNet generalization \citep{Zhang2021}.
We further compare with \emph{CFor}, a random forest-based approach for causal estimation \citep{Wager2018}, and 
three propensity score estimators: \emph{PScov} uses the observed covariates, \emph{PSpca} maps them into a $d_z$-dimensional feature vector, and \emph{PSlat} learns an estimator on the latent encoding space $\mcZ$ our approach infers. All three use an ElasticNet \citep{zou2005} to infer the estimator.

We report the following ablations of our model. 
\emph{(Our)} optimizes \eqref{eq:elbo} with an unconditional, i.e., non-identifiable prior~$p(\mbz)$, \emph{(+I)} specifies the conditional $p(\mbz|\mbc)$. 
\emph{(+sep)} infers separate representations $\mcU$ for each group.
For simplicity, all variants follow the TarNet architecture as closely as possible. 
Finally, \emph{(+snet)} and \emph{(+tedvae)} demonstrate that our proposal can easily be implemented with modern deep models by simply replacing the TarNet architecture backbone with SNet or TEDVAE~(see \cref{sec:app-modern} for a more detailed discussion of such replacements). 

\paragraph{Performance Metrics.} 
If outcome information is available for both groups (case \emph{(a)}) we evaluate the root mean squared error (RMSE) between true and estimated CATE values, also known as the \emph{precision in estimation of heterogeneous effects (PEHE)}~\citep{hill2011}. 
Otherwise (case \emph{(b)}) we compute the ATT between the treated and the matched subset of the control and report the absolute error (AE) of the estimate. 
We consider \emph{within-sample} and \emph{out-of-sample} performance. Within-sample considers covariates~$\mbx$ and outcomes $y=Y(t)$ already observed during training and requires inferring the counterfactual $Y(1-t)$. Out-of-sample performance considers previously unseen covariates~$\mbx$ and potential outcomes for the prediction.

\subsection{Full Outcome Observation}  

We summarize the performance of CATE estimation on IHDP covariates assuming outcomes are available for both groups (treatment and control) in \cref{tab:main-joint} (a). 
Comparing the performance of deterministic and latent-variable approaches, we observe that switching from a sufficiently advanced deterministic model (\emph{TNet, TARNet, CFRNet, SNet}) to a simple generative model (\emph{CEVAE}) tends to reduce performance throughout the different scenarios. As was argued already in prior work by \citet{lu2020}, forcing the latent encoding $\mcZ$ to fully encode the covariates instead of focusing on being predictive constraints too much. However, splitting the latent space into $\mcU$ and~$\mcZ$ clearly improves upon the deterministic baselines (\emph{Ours}). If $\mcU$ is sufficiently high-dimensional, further splitting it into group-specific latent spaces provides little to no benefit in this setup (\emph{+sep}). Switching from an unconditional prior ($p(\mbz)$) to a conditional one ($p(\mbz|\mbc)$) improves the predictive performance even further (\emph{+I}). Finally, switching to a more modern architecture (\emph{+snet}) gives an additional reduction in RMSE. The tree-based baseline (\emph{CFor}) performs significantly worse than all neural net-based methods on this setting, as was already observed by \citet{curth2021}, and fails completely if there is only limited overlap between the two sets of covariates.
Further results are provided in \cref{sec:app-full}.

\subsection{Partial Outcome Observation}

We summarize the performance of ATT estimation on IHDP covariates assuming outcomes are available only for the control group in \cref{tab:main-joint} (b).
In this setting, as before, switching from a deterministic baseline to a generative variant requires the added structure provided by a separate latent space $\mcU$ for significant improvements~(\emph{Ours}). 
While relying on an identifiable version of our approach (\emph{+I}) tends to improve performance, it is not significant, while relying on an SNet structure may even decrease performance depending on the setup. Comparing all baselines with a VAE that simply learns a joint representation for both groups without any outcome information (\emph{VAE}) shows that some guidance, even if it is only accessible from a single group, is necessary to infer a representation suitable for subsequent matching~(\emph{Ours}). 
While matching based on a propensity score estimator is not competitive with the deep models, it improves significantly when it is trained on an encoded set of covariates (\emph{PSlat}) compared to the observed ones (\emph{PScov}). 
Further results are provided in \cref{sec:app-partial}.

\subsection{Missingness} \label{sec:missing}
\begin{figure}%
    \centering
    \includegraphics[width=0.98\columnwidth]{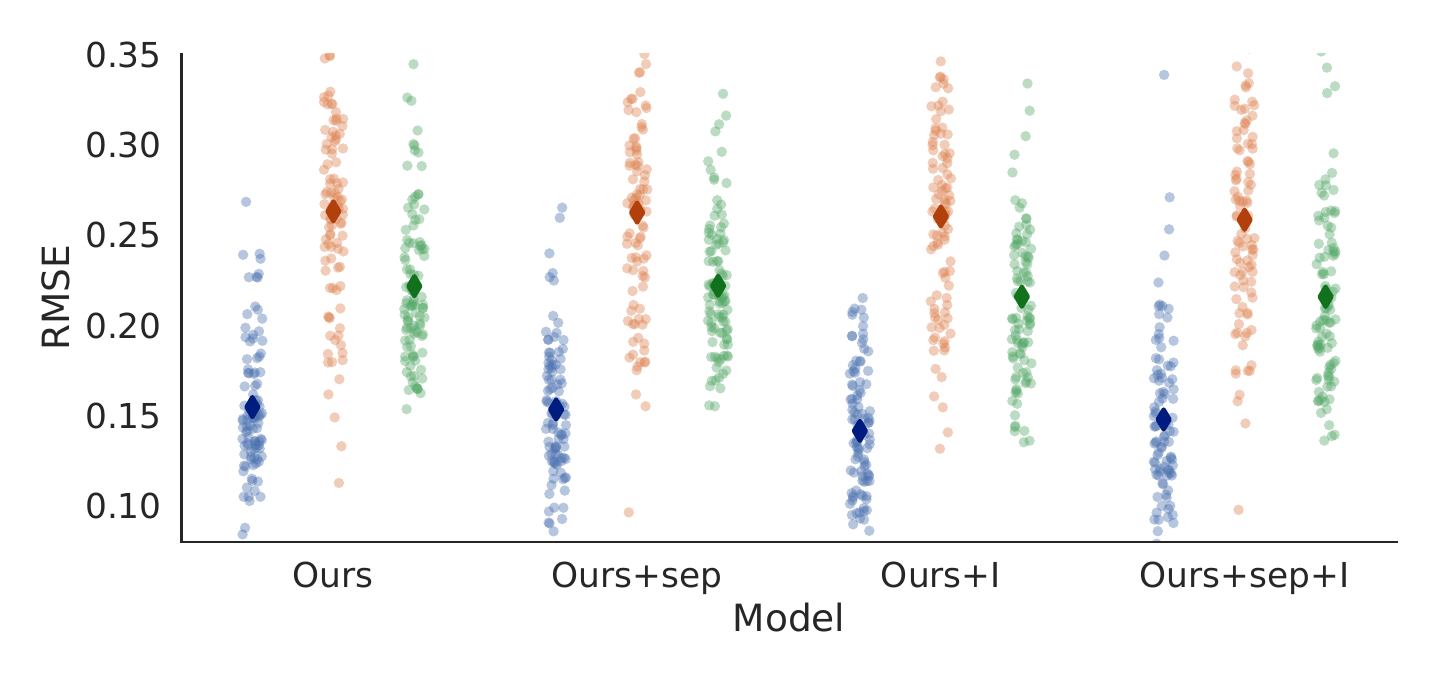}
    \caption{\emph{Missingness.} 
    Moving from a scenario where all covariates are observed ({\color{miss1} $\blacklozenge$}) to one with structured missingness that is not modeled ({\color{miss2} $\blacklozenge$}) reduces performance, as expected. 
    Modeling the MNAR pattern as part of our generative model ({\color{miss3} $\blacklozenge$}) reliably improves performance in all of our variants. Visualized is the RMSE of within-sample CATE estimation in the \emph{all+high} scenario. Shown are 100 random replications along with their respective means ($\blacklozenge$). 
    }\label{fig:missing}
\end{figure}

We summarize the performance of CATE estimation on IHDP covariate in a missing not at random~(MNAR) scenario in \cref{fig:missing}. To simulate this scenario we randomly drop covariates, with the probability depending on the observed values for $\mbx_n$  
Throughout, we assume that missingness only occurs in $\mbx$, not in $\mbc$ or $y$. 
Increasing the amount of missingness in the covariates decreases performance in all models (see \cref{sec:app-missingness}). Proper modeling of the MNAR structure consistently gives significant performance improvements for all variations of our proposal. We discuss the precise mechanism of how we create this data set in  \cref{sec:app-data-ihdp}.
See \cref{sec:app-missingness} for an ATT estimation scenario and further results.

\subsection{Survival} 

\begin{figure*}
    \centering
    \includegraphics[width=0.32\textwidth]{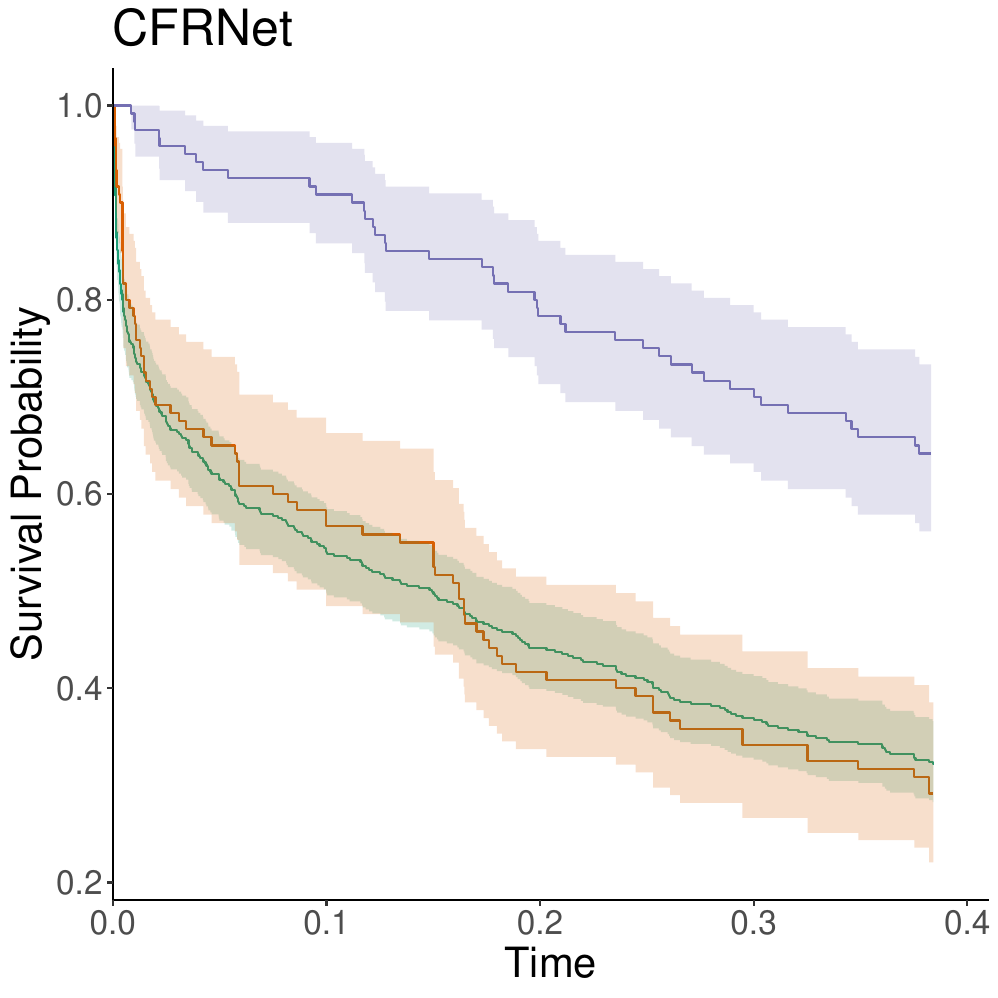}\hspace{0.5em} 
    \includegraphics[width=0.32\textwidth]{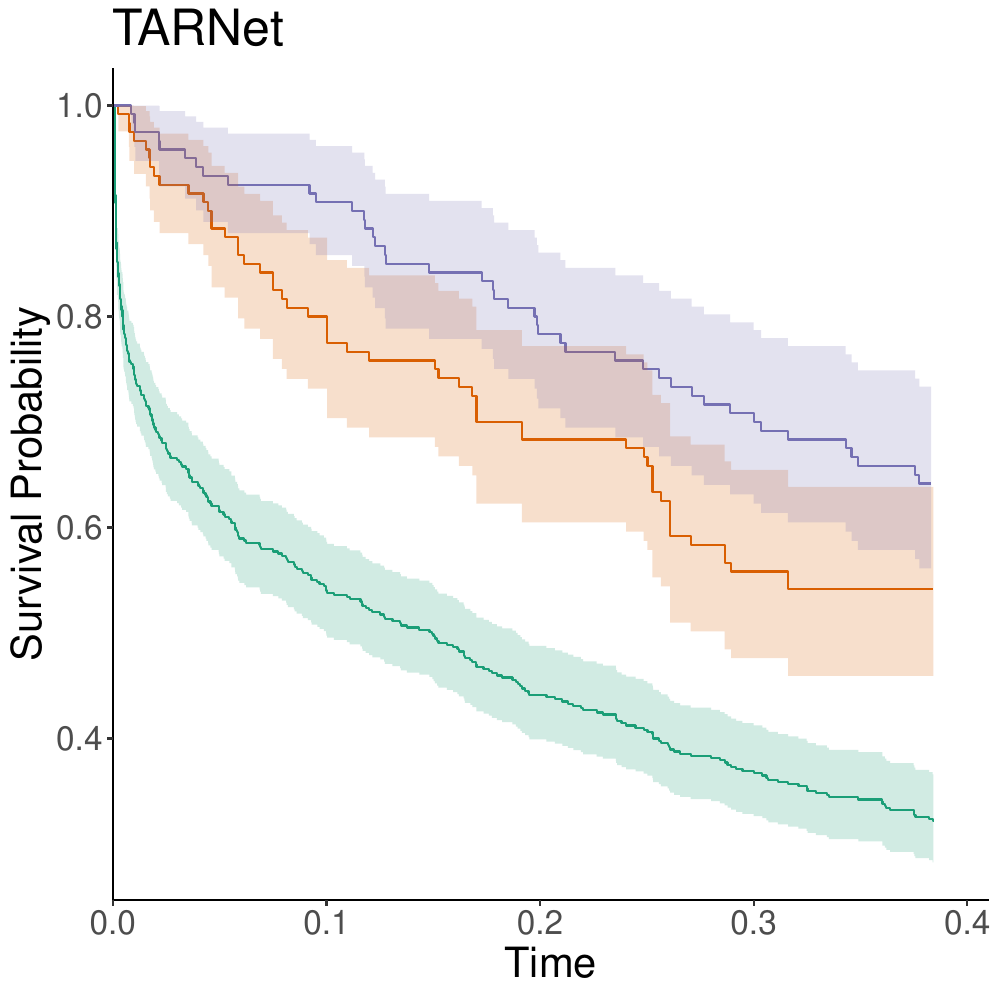}\hspace{0.5em}
    \includegraphics[width=0.32\textwidth]{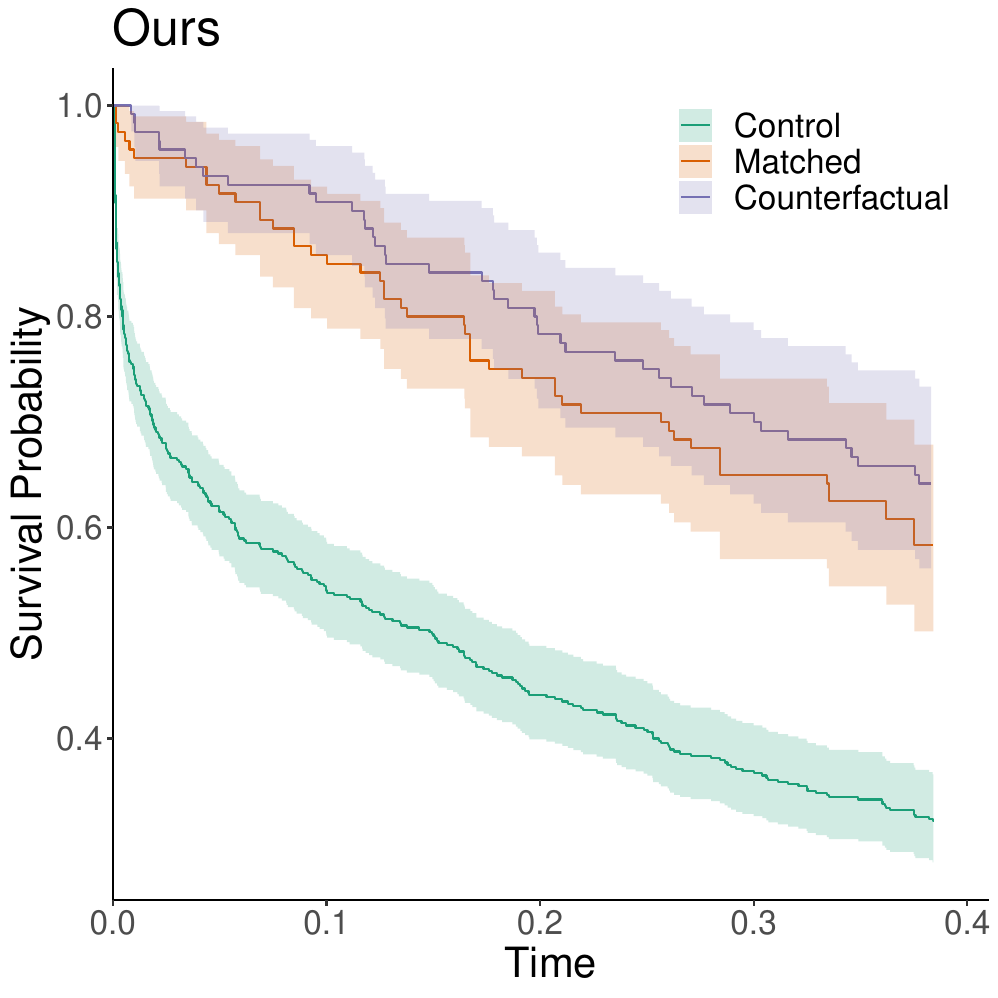}
    \caption{\emph{Survival.} 
    Kaplan-Meier estimates of survival curves for three of the models for a single random seed. During training survival information for the treatment group is unavailable and we observe survival times only for the {\color{survexternal} control group}. After patient matching, the {\color{survmatched} matched subset} is compared against the unknown, {\color{survtrue} counterfactual survival curve} of the single-arm group. 
    The shaded areas are the 95\% confidence intervals on the respective Kaplan-Meier estimators.
    While the match found by \emph{CFRNet} overlaps completely with the unmatched control group, \emph{TARNet} can get close to the desired survival curve. \emph{Ours} largely overlaps with it.  
    }
    \label{fig:survival}
\end{figure*}

So far our evaluation has assumed a scalar outcome $y$, i.e.,~$\mcY = \mdR$. However, our method is not limited to this case.  
We now demonstrate its usefulness in the area of survival analysis \citep{Clark2003}, where $y$ specifies a (potentially censored) survival time. Given access to the survival times of a control group via, e.g., their electronic health records, the goal is to select a suitable subgroup whose survival curve mimics the true, but unknown, counterfactual survival curve of the single-arm treatment group. 
We create synthetic survival data based on IHDP covariates and assume a Weibull distribution for $p(y|t,\mbz)$, both for our model as well as the baselines. 
To evaluate the matching performance, we evaluate the squared error between the estimated hazard ratios relying either on the true counterfactual or the matched subset. Our proposal method improves upon the deterministic as well as generative baselines.
See \cref{fig:survival} for an exemplary visualization on a subset of the methods and \cref{tab:survival} for a summary of the results.

\begin{table}
    \centering
    \caption{\emph{Survival.} Median squared error between the estimated and true hazard ratios for survial curve estimation. Our approach improves both upon its deterministic as well as its generative baselines in most variants. 
    }
    \label{tab:survival}
    \adjustbox{max width=0.99\columnwidth}{
    \begin{tabular}{lcc}
    \toprule
\textsc{Method} & Median Error\\
\midrule
    Naive & $0.394\sd{~[0.093,0.898]}$ \\    \midrule
    TARNet  & $0.162 \sd{~[0.018,0.501]}$\\
    CFRNet & $0.513 \sd{~[0.166,1.059]}$\\
    SNet & $0.486 \sd{~[0.173,1.120]}$\\
    VAE & $0.288 \sd{~[0.064,0.663]}$\\
    CEVAE  & $0.173 \sd{~[0.022,0.572]}$\\
    \midrule
    Ours   & $\mathbf{0.135 \sd{~[0.053,0.526]}}$\\
    \quad+I   & $0.180 \sd{~[0.020,0.448]}$\\
    \quad+sep & $0.155 \sd{~[0.036,0.509]}$\\
    \quad+sep+I & $0.156 \sd{~[0.033,0.473]}$\\
    \quad+snet & $0.331 \sd{~[0.060,1.425]}$\\
    \quad+snet+I & $0.201 \sd{~[0.316,1.120]}$\\
         \bottomrule
    \end{tabular}}\\
    \tiny{\emph{median over 100 random replications with [lower, upper quartile];\\ lowest median marked \textbf{bold}}}
\end{table}

\subsection{Real-World Experimental Data} 
\begin{table}%
    \centering
    \caption{\emph{Real-World Medical Data.} Combining single-arm trial data with electronic health records. We observe the same pattern as before, i.e., switching to our generative proposal improves upon the respective baselines.
    }\label{tab:resultsprivate}
    \adjustbox{max width=0.95\columnwidth}{
    \begin{tabular}{lcc}
        \toprule
        \textsc{AE of ATT} & within-sample & out-of-sample \\
        \midrule 
        PScov & $2.984 \sd{\pm1.444}$ &$2.477\sd{\pm0.740}$ \\
        TARNet  & $2.514 \sd{\pm 1.007}$ & $2.236\sd{\pm{0.892}}$ \\
        CFRNet & $2.271 \sd{\pm 0.884}$ & $2.230\sd{\pm 0.865}$\\
        SNet & $3.032 \sd{\pm1.141}$&$2.319\sd{\pm0.678}$\\
        VAE & $2.445\sd{\pm0.828}$ & $2.123\sd{\pm0.886}$\\
        CEVAE  & $2.587\sd{\pm1.051}$ & $2.221\sd{\pm0.954}$\\
        \midrule
        Ours   & $2.143 \sd{\pm 0.918}$ & $2.032 \sd{\pm 0.874}$\\
        \quad+I   & $2.277\sd{\pm 0.935}$ & $\underline{1.997\sd{\pm0.899}}$\\
        \quad+sep & $2.893\sd{\pm1.332}$&$3.163\sd{\pm1.381}$\\
        \quad+sep+I & $2.775\sd{\pm1.266}$&$2.520\sd{\pm1.109}$\\
        \quad+snet & $2.451\sd{\pm0.901}$&$2.277\sd{\pm0.892}$\\
        \quad+snet+I & \underline{$2.121\sd{\pm0.824}$}& $2.247\sd{\pm0.892}$\\
        \bottomrule
    \end{tabular}}\\
    \tiny{\emph{mean $\pm$ standard error over 5 random replications;\\ 
    lowest mean \underline{underlined}}
    }
\end{table}

We summarize the performance of ATT estimation in \cref{tab:resultsprivate}. 
In our clinical data setup, as was the case for IHDP, we cannot evaluate performance against a true treatment outcome, given that the counterfactual reality is necessarily unknowable. 
We therefore have to again rely on synthetic outcomes on top of real covariates. See \cref{sec:app-data} for details on their creation. 
Evaluating the absolute error of ATT estimation, we observe that switching to our proposal improves performance over the baselines. Overall, a conditional prior $p(\mbz|\mbc)$ tends to improve upon their non-identifiable partner methods while splitting $\mcU$ into two separate sub-spaces tends to hurt overall performance.

\section{CONCLUSION}

We contributed with this work a principled latent-variable model with the ability to estimate treatment effects from single-arm trials with external controls.
The model learns a predictive low-dimensional latent space as well as  separate (group-specific) representations to account for the different characteristics between treated and untreated patients.
We performed an extensive ablation study on both synthetic benchmark data as well as a real-world combination of RCT and EHR data.
Our model improved upon an extensive set of baselines across the benchmarks. %
In this work, we made technical choices regarding architectures and hyperparameters such that we can conclude on the importance of the different modeling choices proposed in our approach, without introducing experimental biases. The next step is to fine-tune technical choices, e.g., by incorporating further advanced architectures such as the FlexTENet~\citep{curth2021flex}, or estimators, such as the R-Learner~\citep{lu2020,nie2020} in the same way as we already demonstrated.

\section*{ACKNOWLEDGEMENTS}
This work was supported by the Academy of Finland Flagship programme: Finnish Center for
Artificial Intelligence FCAI. Samuel Kaski was supported by the UKRI Turing AI World-Leading
Researcher Fellowship, [EP/W002973/1]. 
This work was supported by the Research Council of Finland (decision number: 359135).
We acknowledge the computational resources provided by the Aalto Science-IT project.
Additional computations have been performed using resources at HUS Acamedic analytics platform.

\bibliographystyle{plainnat}
\bibliography{references}

\section*{Checklist}

 \begin{enumerate}

 \item For all models and algorithms presented, check if you include:
 \begin{enumerate}
   \item A clear description of the mathematical setting, assumptions, algorithm, and/or model. [Yes]\\
   \textit{See \cref{sec:app-theory}.}
   \item An analysis of the properties and complexity (time, space, sample size) of any algorithm. [Yes]\\
   \textit{See \cref{sec:app-runtime} for an empirical evaluation of the runtime.}
   \item (Optional) Anonymized source code, with specification of all dependencies, including external libraries. [No]\\
   \textit{An implementation of all models included in the experiments will be available upon acceptance.}
 \end{enumerate}

 \item For any theoretical claim, check if you include:
 \begin{enumerate}
   \item Statements of the full set of assumptions of all theoretical results. [Yes]\\
   \textit{See \cref{sec:app-theory}.}
   \item Complete proofs of all theoretical results. [Yes]\\
   \textit{See \cref{sec:app-theory}.}
   \item Clear explanations of any assumptions. [Yes]\\
   \textit{See \cref{sec:app-theory}.}
 \end{enumerate}

 \item For all figures and tables that present empirical results, check if you include:
 \begin{enumerate}
   \item The code, data, and instructions needed to reproduce the main experimental results (either in the supplemental material or as a URL). [Yes]\\
   \textit{See \cref{sec:app-expdetails}.}
   \item All the training details (e.g., data splits, hyperparameters, how they were chosen). [Yes]\\
   \textit{See \cref{sec:app-expdetails}.}
   \item A clear definition of the specific measure or statistics and error bars (e.g., with respect to the random seed after running experiments multiple times). [Yes] \\
   \textit{See \cref{sec:app-expdetails}.}
   \item A description of the computing infrastructure used. (e.g., type of GPUs, internal cluster, or cloud provider). [Yes] \\
   \textit{See \cref{sec:app-expdetails}.}
 \end{enumerate}

 \item If you are using existing assets (e.g., code, data, models) or curating/releasing new assets, check if you include:
 \begin{enumerate}
   \item Citations of the creator If your work uses existing assets. [Yes]
   \item The license information of the assets, if applicable. [Yes]
   \item New assets either in the supplemental material or as a URL, if applicable. [Yes]
   \item Information about consent from data providers/curators. [Yes]
   \item Discussion of sensible content if applicable, e.g., personally identifiable information or offensive content. [Yes]
 \end{enumerate}

 \item If you used crowdsourcing or conducted research with human subjects, check if you include:
 \begin{enumerate}
   \item The full text of instructions given to participants and screenshots. [No]\\
   \textit{This work did not collect new data from human subjects. The real-world data deals with prior data collected from patients. The specifics of the data collection are currently not included to keep the anonymity, but will be available upon publication.}
   \item Descriptions of potential participant risks, with links to Institutional Review Board (IRB) approvals if applicable. [No] \\
   \textit{See above.}
   \item The estimated hourly wage paid to participants and the total amount spent on participant compensation. [Not Applicable]\\
   \textit{Patients were not payed for their treatment.}
 \end{enumerate}

 \end{enumerate}

\onecolumn
\appendix

  \hsize\textwidth
  \linewidth\hsize \toptitlebar {\centering
  {\Large\bfseries Latent-variable Modeling of Treatment Effects\\ from Single-arm Trials \\[1em]
Supplementary Materials \par}}
 \bottomtitlebar \vskip 0.2in plus 0fil minus 0.1in

\section{FURTHER RELATED WORK}\label{app:related}
\paragraph{Multi-task, Multi-modal, and Transfer Learning.} 
How to combine information from multiple data sources is an ongoing research question.
Multi-task learning considers the problem of solving multiple related tasks from a single data set~\citep{caruana1997,maurer2016,makino2022}.
Multi-modal learning is closely related but focuses instead on using heterogeneous data sources for a single target task~\citep{joy2022}. 
Transfer learning considers the task of transferring a learned model from one data set to another~\citep{pan2010,tripuraneni2020}.
Our task is related to this field via domain adaptation~\citep{ben2010}. We assume two different data sources with similar yet distinct characteristics and extract a predictive representation by combining the two.

\section{THEORY}\label{sec:app-theory}

\subsection{Identifiability}\label{sec:app-ident}
\subsubsection{Prior Work}
Throughout this work we follow the approach of \citet{khemakhem2020}, who rely on auxiliary covariates for identifiability guarantees. See also the recent work by \citet{Xi2023}, who extend this work by providing additional results and guarantees.

\citet{khemakhem2020} observe that a general unconstrained latent-variable model of the form
\begin{equation*}
    p(\mbx,\mbz) = p(\mbx|\mbz)p(\mbz),
\end{equation*}
where $\mbx$ is assumed to be observed and $\mbz$ to be latent is, in general, not identifiable in the sense of the implication
\begin{equation*}
    \forall(\theta, \theta'): \quad p_\theta(\mbx)=p_{\theta'}(\mbx) \Rightarrow \theta = \theta'.
\end{equation*}

To solve this problem, they follow prior work by \citet{pmlr-v89-hyvarinen19a} and introduce an additional observed variable~$\mbc$
\begin{equation*}
    p_\theta(\mbx,\mbz|\mbc) = p_{f}(\mbx|\mbz)p_{\mbT,\lambda}(\mbz|\mbc),
\end{equation*}
with $\theta = (f,\mbT,\lambda)$ and $f$ an injective function, e.g., a neural net. The likelihood $p_{f}(\mbx|\mbz)$ is assumed to be of the form
\begin{equation*}
    p_{f}(\mbx|\mbz) = f(\mbz) + \veps,
\end{equation*}
with $\veps \sim p(\veps)$, i.e., composed of an additive independent noise source.
The covariates $\mbx$ are assumed to be continuous.\footnote{\citet{khemakhem2020} speculated that this constraint on a continuous covariate could also be extended to discrete covariates, but \citet{Xi2023} show, that this is not provable in its current form. Nevertheless, both \citet{khemakhem2020} and we do observe good empirical performance nevertheless also for discrete covariates $\mbx$.}
The conditional prior in turn is assumed to be conditionally factorial into a product of exponential family distributions, 
\begin{equation*}
    p_{T,\lambda}(\mbz|\mbc) = \prod_{i=1}^{d_z}\frac{m(z_i)}{Z(\mbc)}\exp\left(\mbT_i(z_i)^\top \lambda(\mbc)\right),
\end{equation*}
where $\mbT_i$ are the sufficient statistics with $\dim(\mbT_i)=k$, and $\lambda(\cdot)$ some function of $\mbc$, e.g., a neural net. Although this factorization assumption is sufficient within our current setup, it has since been shown that it is not necessary \citep{Lu2022,Xi2023}.

We quote the following two definitions and theorems by \citet{khemakhem2020} following their numbering scheme and adapting them to our notation. 

\paragraph{Definition 1.} 
\textit{Let $\sim$ be an equivalence relation on $\Theta$. We say that $p(\mbx|\mbz)p(\mbz)$ is identifiable up to $\sim$ if
\begin{equation*}
    p_\theta(\mbx) = p_{\theta'}(\mbx) \Rightarrow \theta \sim \theta'.
\end{equation*}
The elements of the quotient space $\Theta/_\sim$ are called the identifiability classes.}

\paragraph{Definition 2.} 
\textit{ Let $\sim$ be the equivalence relation on $\Theta$ defined as follows:
\begin{equation*}
    (f,\mbT,\lambda) \sim (\tilde f,\tilde \mbT,\tilde \lambda) \Leftrightarrow \exists A,\mbb \quad \mbT(f^{-1}(\mbx)) = A\tilde\mbT(\tilde f^{-1}(\mbx)) + \mbb \quad \forall \mbx \in \mcX
\end{equation*}
where $A$ is an $d_zk\times d_zk$ matrix and $\mbb$ is a vector. If $A$ is \emph{invertible}, we denote this relation by $\sim_A$. If $A$ is a \emph{block permutation} matrix, we denote it by $\sim_P$.
}

\paragraph{Theorem 1.}
\textit{
Assume that we observe data sampled from a generative model defined according to the model specified above with parameters $(f,\mbT,\lambda)$. Assume the following holds:
\begin{itemize}
    \item[(i)] The set $\{\mbx\in \mcX|\varphi_\veps(\mbx) = 0\}$ has measure zero, where $\varphi_\veps$ is the characteristic function of $p(\veps)$;
    \item[(ii)] the mixing function $f$ is injective;
    \item[(iii)] the sufficient statistics $\mbT$ are differentiable almost everywhere and $(T_{i,j})_{1\leq j\leq k}$ are linearly independent on any subset of $\mcX$ of measure greater than zero;
    \item[(iv)] there exist $d_zk+1$ distinct points $\mbc^0,\ldots,\mbc^{d_zk}$ such that the matrix
    \begin{equation*}
        L = \big(\lambda(\mbc^1) - \lambda(\mbc^0),\ldots,\lambda(\mbc^{d_zk}) - \lambda(\mbc^0)\big),
    \end{equation*}
    of size $d_zk\times d_zk$ is invertible,
\end{itemize}
then the parameters $(f,\mbT,\lambda)$ are $\sim_A$-identifiable.
}

\paragraph{Theorem 4}
\textit{
Assume the following:
\begin{itemize}
    \item[(i)] The family of distributions $q_\phi(\mbz|\mbx,\mbc)$ contains $p_\theta(\mbz|\mbx,\mbc)$,
    \item[(ii)] we maximize 
    \begin{equation*}
        \Ep{\mcD}{\Ep{q_\phi(\mbz|\mbx,\mbu)}{\log p_\theta(\mbx,\mbz|\mbc) - \log q_\phi(\mbz|\mbx,\mbc)}},
    \end{equation*}
    with respect to both $\theta$ and $\phi$,
\end{itemize}
then in the limit of infinite data $\mcD$, the VAE learns the true parameters $\theta^*=(f^*,\mbT^*,\lambda^*)$ up to an equivalence class~$\sim_A$.
}

\subsubsection{Our setup}

The joint probability distribution we use throughout most part of the paper is  
\begin{equation*}
   p(\mbx, y, t, \mbu,\mbz|\mbc) = p(\mbx|\mbu,\mbz)p(t|\mbz)p(y|\mbz)p(\mbu)p(\mbz|\mbc),
\end{equation*}
where for this discussion we assume $\mbx \in \mdR^{d_x}$, i.e., all covariates are continuous and normally distributed. As discussed above, the guarantees do not hold for discrete covariates. Throughout our experiments, we still include them, and empirically still observe good results, but to guarantee identifiability they could also be dropped from the likelihood, and, e.g., just be used as additional inputs to the variational posterior. Similarly, we focus only on the latent-variable model with respect to~$\mbx$, instead of $(\mbx,y)$ as \emph{(i)} $y$ may be partially unobserved and is not part of the variational posterior violating the assumptions of Theorem~4, and \emph{(ii)} $p(\mby|t,\mbz)$ is, e.g., in the survival experiment a Weibull distribution, i.e., does not fulfill the constraint of an independent noise source. 
Note also that Theorem~4 requires us to optimize the original ELBO, i.e., without any regularization terms. We only include these in the two-arm setup but drop them from the one-arm setup to fulfill this constraint.

We keep the prior $p(\mbu)$ unconstrained and only constrain $p(\mbz|\mbc)$ as a factorizing normal distribution
\begin{equation}\label{eq:condprior}
    p(\mbz|\mbc) = \prod_{i=1}^{d_z}m(z_i)/Z(\mbc)\exp\left(T(z_i)^\top\lambda(\mbc)\right),
\end{equation}
with $\lambda(\cdot)$ parameterized by a neural network. Throughout the experiments, we fix the prior variance of $p(\mbz|\mbc)$ and only parameterize the mean, that is $k=1$. As we require $d_zk+1$ distinct values, this allows us to use a lower variety in $\mbc$, at the price of only being able to guarantee $\sim_A$ identifiability, instead of $\sim_P$~\citep[Theorem 3]{khemakhem2020}.

This gives us, after marginalizing over $y,t$ 
\begin{align*}
    p(\mbx,\mbz|\mbc) &= \int p(\mbx|\mbu,\mbz) p(\mbu)p(\mbz|\mbc)\d\mbu\\
    &= \Ep{p(u)}{p(\mbx|\mbu,\mbz)}p(\mbz|\mbc)\\
    &= \Ep{p(u)}{p_\veps(\mbx - f(\mbu,\mbz))} p(\mbz|\mbc)\\
    &= p_\veps\Big(\mbx - \underbrace{\Ep{p(u)}{f(\mbu,\mbz)}}_{\deq h(\mbz)}\Big)p(\mbz|\mbc),
\end{align*}
where $p_\veps(\mbx - h(\mbz)) \deq p_\veps(\veps) = \Norm(\veps|0,\1)$. Assuming the remaining conditions of Theorem~1 hold, this gives us the desired~$\sim_A$ identifiability for $(h,\mbT,\lambda)$.

\subsection{Missingness}\label{sec:app-missing}

Not only can missing covariate information be assumed to be present in most real-world data sets, e.g., electronic health records, but one should also expect it to be missing not at random (MNAR)~\citep{rubin2005}, i.e., the missingness pattern depends on the missing covariates themselves. 
To properly account for that we extend our model with binary masking variables~${\mbm \in \mcM = \{0,1\}^{d_x}}$, where $m_{ni}=0$ indicates that the $i$-th covariate of observation $n$ is missing. We observe 
\begin{equation*}
    \tilde \mbx = \mbm \circ\mbx + (1 - \mbm)\circ \boldsymbol{\eta},
\end{equation*}
where $\circ$ is elementwise multiplication, and $\boldsymbol{\eta}$ some imputation scheme, e.g., $\boldsymbol{\eta} = \boldsymbol{0}$ for zero imputation.

\begin{figure}
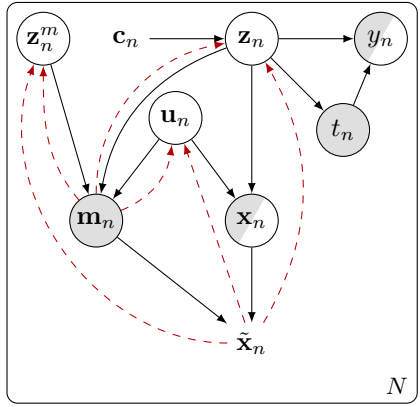

\begin{center}
   \tikz{
       \colorlet{infer}{red!70!black}
        \node[latent] (z) {$\mbz_n$};
        \node[left=of z] (c) {$\mbc_n$};
        \node[semifill={gray!25, ang=60}, draw=black,inner sep=1pt, minimum size=20pt, font=\fontsize{10}{10}\selectfont, node distance=1, below=of z, yshift=-2em, xshift=0em] (x) {$\mbx_n$};
        \node[below=of x] (tx) {$\tilde \mbx_n$};
        \node[semifill={gray!25, ang=60}, draw=black,inner sep=1pt, minimum size=20pt, font=\fontsize{10}{10}\selectfont, node distance=1, right=of z] (y) {$y_n$};
        \node[latent, left=of z, yshift=-3em, xshift=2em] (u) {$\mbu_{n}$};
        \node[obs, left=of x, xshift=-1em] (m) {$\mbm_{n}$};
        \node[latent, left=of z, xshift=-3em] (zm) {$\mbz_{n}^m$};
        \node[obs, below right=of z] (t) {$t_n$};
        \draw[->] (z) to (x); 
        \draw[->] (z) to (t); 
        \draw[->] (z) to (y); 
        \draw[->] (zm) to (m); 
        \draw[->] (u) to (x); 
        \draw[->] (u) to (m); 
        \draw[->] (m) to (tx); 
        \draw[->] (t) to (y); 
        \draw[->] (x) to (tx); 

        \draw[->] (c) to (z);

        \draw[->, dashed,infer] (tx) to[bend right] (z);
        \draw[->, dashed,infer] (tx) to (u);
        \draw[->, dashed,infer] (tx) to[out=180, in=-110] (zm);
        \draw[->, dashed,infer] (m) to[out=130, in=-90] (zm);
        \draw[->, dashed,infer] (m) to[bend right] (u);
        \draw[->, dashed,infer] (m) to[out=90, in=190] (z);
        \draw[->] (z) to[out=200, in=80] (m);
        \plate{}{(m)(tx)(zm)(u)(x)(y)(t)(z)}{$N$};
   }
 
\end{center}
\caption{\emph{Missing not at random model.} Empty, partially filled, and filled circles represent latent, partially observed and observed variables. $\tilde \mbx$ is deterministic as described in the  text.
}\label{fig:app-missing}
\end{figure}

We model MNAR by assuming a prior over $\mbm$ that depends on $\mbu$ and $\mbz$, to account for the not-at-random structure, and assume an additional latent $\mbz^m$ to model any remaining pattern. 
Our joint model is given as
\begin{align*}
    \mbu &\sim p(\mbu) = \Norm(\mbu|0, \sigma_u^2\1),\\
    \mbz|\mbc &\sim p(\mbz|\mbc) = \Norm\left(\mbz|\lambda(\mbc), \sigma_{z}^2\1\right),\\
    \mbz^m &\sim p(\mbz^m) = \Norm(\mbz|0, \sigma_{z^m}^2\1),\\
    \mbm|\mbu, \mbz, \mbz^m &\sim \prod_{i=1}^{d_x}\Ber\big(m_i|\sigma(g(\mbz^m, \mbu, \mbz)_i)\big),\\
    \mbx|\mbm, \mbu,\mbz &\sim p(\mbx|\mbm, \mbu,\mbz),\\
    t &\sim p(t|\mbz) = \Ber\big(t|\sigma(f(\mbz))\big),\\
    y &\sim p(y|t, \mbz)= \Norm\big(y|\mu^y_t(\mbz), \alpha^{-1}\big),
\end{align*}
where $f(\cdot)$, $g(\cdot)$, $\mu_t^y(\cdot)$ are neural nets, and $\sigma(x) \deq 1/(1 + \exp(-x))$ is the logistic sigmoid. 
As in the main text, we only focus on identifiability with respect to $\mbz$.
We assume $p(\mbx|\mbm, \mbu,\mbz)$ to factorize over the covariates, with each covariate likelihood being modeled via normal, Bernoulli, or categorical distributions depending on their respective domains.
See \cref{fig:app-missing} for a summarizing plate diagram. 

We define $\mbU = (\mbu_1,\ldots,\mbu_N)$ and $\mbZ^m$, $\mbZ$, $\mbX$ are defined analogously. We again rely on a mean-field variational posterior, given as 
\begin{align*}
q(\mbU,\mbZ^m,\mbZ|\mbC,\mbX) &= \prod_{n=1}^N q(\mbu_{n}|\mbm_n,\tilde\mbx_n)q(\mbz_n|\mbc_n,\mbm_n, \tilde\mbx_n)q(\mbz_n^m|\mbm_n, \tilde \mbx_n)\\
&=\prod_{n=1}^N \Norm\big(\mbu_{n}|\mu^u(\mbm_n,\tilde \mbx_n), \sigma^u(\mbm_n,\tilde \mbx_n)\big)\Norm\big(\mbz_n|\mu^{z}(\mbc_n,\mbm_n,\tilde\mbx_n), \sigma^{z}(\mbc_n,\mbm_n,\tilde\mbx_n)\big)\\
&\hspace{4em}\cdot\Norm\big(\mbz^m_n|\mu^{z^m}(\mbm_n, \tilde\mbx_n), \sigma^{z^m}(\mbm_n, \tilde\mbx_n)\big),
\end{align*}
The full ELBO is given as 
\begin{align*}
    \log~&p(\mcD) \geq \sum_{n=1}^N \Ep{q(\mbu_{n},\mbz_n, \mbz_n^m|\mbc_n, \mbx_n)}{\log p(\mbx_n|\mbm_n,\mbu_{n},\mbz_n) + \log p(\mbm|\mbu, \mbz^m,\mbz) \big.} \\
    &\quad+ \Ep{q(\mbz_n|\mbc_n, \mbx_n)}{\log p(y_n|t_n,\mbz_n) + \log p(t_n|\mbz_n)\big.}\\
                 &\quad -\KL{q(\mbu_{n}|\tilde{\mbx}_n)\big.}{p(\mbu_{n})} - \KL{q(\mbz_n | \mbc_n,\tilde{\mbx}_n)\big.}{p(\mbz_n|\mbc_n)}\\
                 &\quad -\KL{q(\mbz^m_{n}|\mbm_n,\tilde{\mbx}_n)\big.}{p(\mbz^m_{n})}
\end{align*}
See \cref{sec:app-missingness} for related experiments.

\subsection{A split $\mcU$}
The main paper discusses a separate latent space $\mcU$ and formulates the joint model as 
\begin{align*}
    \mbu &\sim p(\mbu) = \Norm(\mbu|0, \sigma_u^2\1),\\
    \mbz &\sim p(\mbz|\mbc) = \Norm(\mbz|\lambda(\mbc), \sigma_z^2\1),\\
    \mbx &\sim p(\mbx|\mbu, \mbz),\\
    t &\sim p(t|\mbz) = \Ber\big(t|\sigma(f(\mbz))\big),\\
    y &\sim p(y|t, \mbz)= \Norm\big(y|\mu^y_t(\mbz), \alpha^{-1}\big).
\end{align*}
Assuming data from $n_\text{data}$ sources, another approach is to use a separate $\mcU$ for each of them, i.e., we have~${\mcU = \mcU_1\times \cdots\times \mcU_{n_\text{data}}}$. The joint then becomes
\begin{align*}
    \mbu^g &\sim p(\mbu^g) = \Norm(\mbu^g|0, \sigma_u^2\1), \quad\text{for $g=1,\ldots, n_\text{data}$},\\
    \mbz &\sim p(\mbz|\mbc) = \Norm(\mbz|\lambda(\mbc), \sigma_z^2\1),\\
    \mbx &\sim p(\mbx|\mbu, \mbz),\\
    t &\sim p(t|\mbz) = \Ber\big(t|\sigma(f(\mbz))\big),\\
    y &\sim p(y|t, \mbz)= \Norm\big(y|\mu^y_t(\mbz), \alpha^{-1}\big).
\end{align*}
In our specific setup of one treatment and one control group, we have $n_\text{data}=2$. However, this can directly be generalized to multiple control sources and separate treatment trials.

\subsection{Outcome only for the control group}
A common assumption in the machine learning literature is to predict treatment effects given pre- and post-treatment information, i.e., covariates as well as outcomes for all treatment and control groups. The additional alternative setting that we consider is that the model is blinded to post-treatment information from the treatment group, i.e., the model has to infer a suitable latent embedding solely from pre-treatment covariates of the single-arm trial as well as covariate and outcome information from the control electronic health records. 

This blinding ensures, that before having selected a fixed set of controls to compare against no post-treatment information is leaked that could potentially bias future tests.

Assuming that we have $N_0$ control observations and $N_1$ treated observations, our data set becomes ${\mcD = \{(x_1,y_1,t_1,),\ldots, (x_{N_0},y_{N_0},t_{N_0}), (x_{N_0+1},t_{N_0+1}), \ldots, (x_{N_0+N_1},t_{N_0 + N_1})\}}$ and the ELBO becomes
\begin{align*}
    \log p(\mcD) &\geq \sum_{n=1}^{N_0} \Ep{q(\mbu_{n},\mbz_n|\mbc_n, \mbx_n)}{\log p(\mbx_n|\mbu_{n},\mbz_n)}+ \Ep{q(\mbz_n|\mbc_n, \mbx_n)}{\log p(y_n|t_n,\mbz_n) + \log p(t_n|\mbz_n)} \\
    &\qquad+ \sum_{n=N_0+1}^{N_0+N_1} \Ep{q(\mbu_{n},\mbz_n|\mbc_n, \mbx_n)}{\log p(\mbx_n|\mbu_{n},\mbz_n)}+ \Ep{q(\mbz_n|\mbc_n, \mbx_n)}{\log p(t_n|\mbz_n)}\\
                 &\qquad -\KL{q(\mbu_{n}|\mbx_n)\big.}{p(\mbu_{n})} - \KL{q(\mbz_n | \mbc_n,\mbx_n)\big.}{p(\mbz_n|\mbc_n)}\\
    &= \sum_{n=1}^{N} \Ep{q(\mbu_{n},\mbz_n|\mbc_n, \mbx_n)}{\log p(\mbx_n|\mbu_{n},\mbz_n)}+ \Ep{q(\mbz_n|\mbc_n, \mbx_n)}{(1 - t_n)\log p(y_n|t_n,\mbz_n) + \log p(t_n|\mbz_n)} \\
                 &\qquad -\KL{q(\mbu_{n}|\mbx_n)\big.}{p(\mbu_{n})} - \KL{q(\mbz_n | \mbc_n,\mbx_n)\big.}{p(\mbz_n|\mbc_n)},
\end{align*}
where we use $N=N_0+N_1$ and recover the original ELBO in \eqref{eq:elbo} up to the correction factor for notational simplicity.

\section{EXPERIMENTAL DETAILS}\label{sec:app-expdetails}
This section goes through various details on the experimental setups. A reference implementation is provided at \texttt{anonymous}.

\subsection{Significance Testing}
Throughout our experiments, we rely on paired Student t-tests to evaluate which models have a comparable performance. The t-tests are computed relatively to the model with the best empirical mean and are one-sided, i.e., whether the alternative has a lower mean. We mark all methods that do not lead to a rejection of the null hypothesis at a significance level of $p < 0.05$ in bold. This applies to all experiments throughout this paper. The lowest empirical mean is additionally underlined. The number of repetitions for each experiment is specified below the respective table of results.  

\subsection{Models Under Consideration}\label{app:baselines}
Within this paper, we compare against a range of deterministic and probabilistic baselines as well as several baselines. We discuss their relation to our method within this section. Unless noted otherwise, we reimplemented all baselines in PyTorch~\citep{NEURIPS2019_9015}.

\paragraph{Non neural net-based baselines.}
\begin{itemize}
    \item \emph{CFor} is a popular causal forest-based approach for causal estimation introduced by \citet{Wager2018} and serves as our main non-neural network-based baseline. Given its lack of a representation space to perform matching on, it can only be used in the scenario where outcome information is available for both groups. We rely on the implementation provided by the EconML package~\citep{econml}.
    \item \emph{PScov}, \emph{PSpca}, and \emph{PSlat} serve as propensity score (PS)-based matching methods that are constructed on various sets of inputs. \emph{PScov} uses the features provided by the original covariates in $\mcX$. \emph{PSpca} first maps them into a $\dim(z)$ space via a principal component analysis (PCA). \emph{PSlat} finally relies on $\mcZ$ space encodings that were provided by one of our neural network-based approaches. Throughout the experiments, we always use the encoding of the best-performing neural net as input to \emph{PSlat}. Independent of their input domain, all three construct their estimator of $\pi(x) = \mdP(T=1|X=x)$ via an ElasticNet~\citep{zou2005}. 
    As these propenstiy score methods do not infer estimators $\mu_t(\cdot)$ we only compare against them in the matching experiments.
    We rely on the ElasticNet implementation provided by scikit-learn~\citep{2020SciPy-NMeth}. 
\end{itemize}

\paragraph{Deterministic neural nets.}
\begin{itemize}
    \item \emph{SingleNet} is a simple neural net, that learns a simple estimator $\mu(\mbx,t)$ by concatenating $t$ to the features $\mbx$, such that $\mu_t(\mbx) = \mu(\mbx,t)$. Its objective is then given as a simple MSE loss,
    \begin{equation*}
        \min \sum_{n=1}^N (\mu(\mbx_n,t_n) - y_n)^2.
    \end{equation*}
    The estimator $\mu(\cdot,\cdot)$ is parameterized by a neural net.
    As it does not infer a shared latent space we only use it in scenario \emph{(a)} where outcome information is available for both arms.
    \item \emph{TNet} slightly generalizes upon \emph{SingleNet} by learning two separate neural net estimators $\mu_0(\mbx)$ and $\mu_1(\mbx)$. Its objective is again an MSE loss
    \begin{equation*}
        \min \sum_{n=1}^N (\mu_{t_n}(\mbx_n) - y_n)^2.
    \end{equation*}
    We parameterize $\mu_0(\cdot)$ and $\mu_1(\cdot)$ by neural nets.
    Similar to the \emph{SingleNet} it does not learn a shared latent space to perform matching on and is therefore not applicable to the matching scenario.
    \item \emph{TARNet} \citep{shalit2017} is a generalization of the \emph{TNet} that consists of three neural nets. $h(\cdot)$ maps the observed covariates into a shared representation space. $\mu_0(\cdot)$ and $\mu_1(\cdot)$ are then estimated based on this representation. The MSE loss is given as
    \begin{equation*}
        \min \sum_{n=1}^N \left(\mu_{t_n}(h(\mbx_n)) - y_n\right)^2.
    \end{equation*}
    \item \emph{CFRNet} \citep{shalit2017} extends \emph{TARNet} with an additional MMD (see \cref{sec:app-regularize} for details) term in the loss for further regularization. 
    \item \emph{SNet} \citep{curth2021} further generalizes upon these by replacing the single $h(\cdot)$ into five sub-nets~$h_i(\cdot), i \in \{1,\ldots,5\}$ that separate the representation space into various sub-spaces, which serve as input to $\mu_t(\cdot)$ as well as a separate classifier $g(\cdot)$ that predicts $t$ given these representations which is optimized via a cross-entropy loss. 
    These subspaces in turn serve as \emph{(i)} input only to $\mu_0$, \emph{(ii)} input to both $\mu_0$, $\mu_1$, \emph{(iii)} input only to $\mu_1$, \emph{(iv)} input to all estimators $\mu_0$, $\mu_1$, $g$, or \emph{(v)} input only to $g$.  In addition, SNet replaces the MMD regularizer with an orthogonalization regularizer on the representation space mappings $h_i(\cdot)$. See \citet{curth2021} for details.
\end{itemize}

Comparing the loss terms of these deterministic models to our ELBO objective \eqref{eq:elbo} they can be interpreted as optimizing (up to variations in the model structure and additional loss terms), 
\begin{equation*}
    \sum_{n=1}^N\Ep{q(\mbz_n|\mbx_n)}{\log p(y_n|\mbz_n,t_n)},
\end{equation*}
for a deterministic delta distribution $q(\mbz|\mbx) = \delta(\mbz - h(\mbx))$, and $p(y|\mbz,t) = \Norm(y| \mu_t(\mbz), 0.5)$.\footnote{The $0.5$ is due to recover the MSE loss without the scaling factor $1/2$ the normal distribution introduces in its exponential.} \emph{SNet} comes even closer by adding an additional $\log p(t|\mbz)$ to the objective.

\paragraph{Generative baselines.} 
While our deterministic baselines only consider the covariates as input features, the generative approaches also rely on learning representations that can (approximately) reconstruct them. All generative models rely on mean-field variational posteriors and use the same prior and likelihood assumptions as far as possible. 

\begin{itemize}
    \item \emph{VAE} is a simple variational auto-encoder \citep{kingma2014}, whose ELBO is given as
    \begin{equation*}
        \sum_{n=1}^N\Ep{q(\mbz_n|\mbx_n)}{\log p(\mbx_n|\mbz_n)} - \KL{q(\mbz_n|\mbx_n)}{p(\mbz_n)}.
    \end{equation*}
    As it lacks a model for the outcome likelihood $y$, we only compare against it in the matching scenarios.
    \item \emph{CEVAE} \citep{louizos2017} provides an output likelihood for the outcome $y$ together with an additional likelihood for $t$, i.e., optimizes an ELBO given as 
    \begin{equation*}
        \sum_{n=1}^N\Ep{q(\mbz_n|\mbx_n)}{\log p(\mbx_n|\mbz_n) + \log p(y_n|\mbz_n,t_n) + \log p(t_n|\mbz_n)} - \KL{q(\mbz_n|\mbx_n)}{p(\mbz_n)}.
    \end{equation*}
    Note that \citet{louizos2017} include additional reconstructive terms in their objective which we leave out to keep a valid ELBO. 
    \item \emph{TEDVAE} \citep{Zhang2021} generalizes \emph{CEVAE} similar to how \emph{SNet} generalized the \emph{CFRNet}. It splits the latent $\mcZ$ into three parts such that the joint is given as 
    \begin{equation*}
    p(\mbx,\mbz^1, \mbz^2,\mbz^3,y,t) = p(\mbx|\mbz^1,\mbz^2,\mbz^3) p(t|\mbz^1,\mbz^2)p(y|\mbz^2,\mbz^3)p(\mbz^1)p(\mbz^2)p(\mbz^3).
    \end{equation*}
    The corresponding ELBO objective is then given as 
    \begin{equation*}
        \sum_{n=1}^N\Ep{q(\mbz_n|\mbx_n)}{\log p(\mbx_n|\mbz_n) + \log p(y_n|\mbz_n^2,\mbz_n^3,t_n) + \log p(t_n|\mbz_n^1,\mbz_n^2)} - \KL{q(\mbz_n|\mbx_n)}{p(\mbz_n)}.
    \end{equation*}
    where we use $\mbz_n = (\mbz_n^1,\mbz_n^2,\mbz_n^3)$ in the notation.
\end{itemize}

\paragraph{Variations of our proposal.}
Throughout the experiments, we evaluate a wide range of variations on our model. We describe each of the building blocks in turn.

\begin{itemize}
    \item \emph{Ours} is the basic model we rely on with a joint given as $p(t,\mbu,\mbx,y,\mbz) = p(t|\mbz)p(\mbu)p(\mbx|\mbu,\mbz)p(y|\mbz,t)p(\mbz)$. The ELBO is given as 
\begin{align*}
    &\sum_{n=1}^N \Ep{q(\mbu_{n},\mbz_n|\mbx_n)}{\log p(\mbx_n|\mbu_{n},\mbz_n)}
    + \Ep{q(\mbz_n|\mbx_n)}{{\log p(y_n|t_n,\mbz_n)} + \log p(t_n|\mbz_n)\big.} \\
                 &\qquad-\KL{q(\mbu_{n}|\mbx_n)\big.}{p(\mbu_{n})}
                 - \KL{q(\mbz_n | \mbx_n)\big.}{p(\mbz_n)} 
\end{align*}
    \item \emph{``+I''} indicates a variation that is identifiable in $\mbz$. Its ELBO differs from \emph{Ours} in the conditional prior and the additional covariates $\mbc$, highlighted in red,
\begin{align*}
    &\sum_{n=1}^N \Ep{q(\mbu_{n},\mbz_n|{\color{red!60!black} \mbc_n,}\mbx_n)}{\log p(\mbx_n|\mbu_{n},\mbz_n)}
    + \Ep{q(\mbz_n|{\color{red!60!black} \mbc_n,}\mbx_n)}{{\log p(y_n|t_n,\mbz_n)} + \log p(t_n|\mbz_n)\big.} \\
                 &\qquad-\KL{q(\mbu_{n}|\mbx_n)\big.}{p(\mbu_{n})}
                 - \KL{q(\mbz_n | {\color{red!60!black} \mbc_n,}\mbx_n)\big.}{p(\mbz_n {\color{red!60!black}| \mbc_n}))} 
\end{align*}
    \item \emph{``+sep''} indicates a variation that splits $\mcU$ into two parts, each responsible for one of the two groups. The differences in its ELBO compared to \emph{Ours} are highlighted in red
\begin{align*}
    &\sum_{n=1}^N \Ep{q({\color{red!60!black} \mbu_n^{t_n}},\mbz_n| \mbx_n)}{\log p(\mbx_n|{\color{red!60!black}\mbu_{n}^{t_n}},\mbz_n)}
    + \Ep{q(\mbz_n|\mbx_n)}{{\log p(y_n|t_n,\mbz_n)} + \log p(t_n|\mbz_n)\big.} \\
                 &\qquad-\KL{q({\color{red!60!black} \mbu_{n}^{t_n}}|\mbx_n)\big.}{p(\mbu_{n})}
                 - \KL{q(\mbz_n |\mbx_n)\big.}{p(\mbz_n)}.
\end{align*}
    \item \emph{``+snet''} uses the deterministic \emph{SNet} structure and regularization, adding a generator $p(\mbx|\mbu,\mbz)$ on top of the representation $\mbz$, where $\mbz$ is the combination of all five subspaces. 
    \item \emph{``+tedvae''} uses the geneative \emph{TEDVAE} structure, adds a separate $p(\mbu)$, and adapts the covariate log-likelihood to $p(\mbx|\mbu,\mbz^1,\mbz^2,\mbz^3)$.
\end{itemize}

We parameterize the probability densities for the model variations as 
\begin{align*}
    p(\mbu),p(\mbu^{t}), p(\mbz), p(\mbz^m) &= \Norm(0, \sigma^2\1),\qquad \text{$\sigma = 1$ for all experiments},\\
    p(\mbz|\mbc) & = \Norm(\lambda(\mbc), \sigma_z^2\1),\qquad \text{$\sigma_z = 1$ for all experiments},\\
    p(x|\mbz) &= \begin{cases}
   \Norm\big(f_{z\to x}(\mbz),\sigma_x\big),   &\text{if $x$ is continuous} \\
    \Ber\big(\sigma(f_{z\to x}(\mbz))\big), &\text{if $x$ is binary} \\
    \Cat\big(\zeta(f_{z\to x}(\mbz))\big), &\text{if $x$ is categorical} 
    \end{cases},\\
    p(x|\mbu,\mbz) &\text{ is defined analogously but with $f_{uz\to x}(f_{u\to uz}(\mbu),f_{z\to uz}(\mbz))$},\\
    p(t|\mbz) &= \Ber\big(\sigma(f_{z\to t}(\mbz))\big),\\
    p(y|\mbz) &= \begin{cases}
        \Norm(f_{z\to y}(\mbz), \sigma_y) &\text{if continuous}\\
        \text{see \cref{app:survival}}  &\text{if time to event}
    \end{cases},\\
    q(\mbu|\mbx) &= \Norm(\mu^u(\mbx),\sigma^u(\mbx)),\\
    q(\mbu^t|\mbx) &= \Norm(\mu^u_t(\mbx),\sigma^u_t(\mbx)),\\
    q(\mbz|\mbx) &= \Norm(\mu^z(\mbx),\sigma^z(\mbx)),\\
    q(\mbz^m|\mbm) &= \Norm(\mu^{z_m}(\mbx),\sigma^{z_m}(\mbx)),\\
    &\hspace{2em}\text{$q(\mbz|\mbc,\mbx)$ is defined analogously with an additional input $\mbc$},\\
    &\hspace{2em}\text{the split $\mbz$ cases, i.e., \emph{+snet}, \emph{+tedvae} split the output of $\mu^z,\sigma^z$ accordingly},
\end{align*}
where $\sigma(\cdot)$ is the logistic sigmoid, $\zeta(\cdot)$ the softmax, and all remaining functions are parameterized via neural networks. Standard deviations $\sigma_x$ and $\sigma_y$ are treated as free parameters that are optimized via gradient descent as part of the training routine.

\subsection{Modern}\label{sec:app-modern}
\emph{TEDVAE}~\citep{Zhang2021} and \emph{SNet}~\citep{curth2021} serve as two examples of more modern generative and deterministic models and how they can easily be adapted to our setup. \emph{SNet} stands in as an example of a deterministic neural net with a more advanced architectural structure and regularization method. Simply interpreting its mapping from $\mcX$ to the representation space $\mcZ$ as the parameterization of an amortized variational posterior and adding a generator that mirrors it by learning a mapping from $\mcZ$ to $\mcX$, as well as adding a separate representation $\mcU$ are all that is needed.
\emph{TEDVAE} provides the same illustration for generative models, whose adaptation is even simpler. Here an encoder-decoder structure is already provided lacking simply a separate $\mcU$ representation. 
Other novel approaches can be adapted to our setup in the same manner. For comparison, we replicate the corresponding rows from \cref{tab:main-joint} below. Our adaptations improve upon these baselines in all but one setting (highlighted in the table via underlining).

\begin{center}
        \adjustbox{max width=0.98\textwidth}{
    \begin{tabular}{lcccccccccc}
    \toprule
    &\multicolumn{4}{c}{(a) full outcome observation (RMSE of CATE)} & \multicolumn{4}{c}{(b) partial outcome observation (AE of ATT)}\\ 
    \cmidrule(lr){2-5} \cmidrule(lr){6-9}
    &\multicolumn{2}{c}{all+high}  & \multicolumn{2}{c}{subset+low}&\multicolumn{2}{c}{all+high}  & \multicolumn{2}{c}{subset+low}  \\
    \cmidrule(lr){2-3}\cmidrule(lr){4-5}\cmidrule(lr){6-7}\cmidrule(lr){8-9}
\textsc{Method} & within sample & out-of-sample & within sample& out-of-sample& within sample & out-of-sample & within sample& out-of-sample \\
    \midrule
    SNet & $0.168 \sd{\pm 0.003}$ &$0.264 \sd{\pm 0.008}$  &$0.211 \sd{\pm 0.008}$ &$\underline{{0.243 \sd{\pm 0.016}}}$& $0.043 \sd{\pm0.003}$ &$0.137 \sd{\pm0.007}$  &$0.221 \sd{\pm0.015}$ & $0.366 \sd{\pm0.026}$\\
    Ours+snet & $0.141 \sd{\pm 0.002}$ &$0.256 \sd{\pm 0.008}$ &${0.185 \sd{\pm 0.007}}$ &${0.253 \sd{\pm 0.017}}$ & ${0.038 \sd{\pm0.002}}$ &${0.117 \sd{\pm0.007}}$ &$0.194 \sd{\pm0.014}$ &$0.322 \sd{\pm0.020}$ &\\
    \midrule
    TEDVAE & $0.176 \sd{\pm 0.002}$ &$0.283 \sd{\pm 0.007}$ &$0.293 \sd{\pm 0.009}$ &$0.336 \sd{\pm 0.018}$ &$0.052 \sd{\pm0.003}$ &$0.179 \sd{\pm0.009}$ &$0.144 \sd{\pm0.009}$ & $0.272 \sd{\pm0.016}$ &\\
    Ours+tedvae & $0.143 \sd{\pm 0.002}$ &$0.257 \sd{\pm 0.008}$ &${0.187 \sd{\pm 0.007}}$ &$0.258 \sd{\pm 0.019}$ &${0.039 \sd{\pm0.003}}$ &${0.119 \sd{\pm0.007}}$ &${0.103 \sd{\pm0.008}}$ &${0.189 \sd{\pm0.015}}$ &\\
    \bottomrule
    \end{tabular}}\\
\end{center}

\subsection{Architectures and Hyperparameters}\label{sec:app-arch_hyper}
We rely on the same set of architectures and hyperparameters throughout the various experiments and methods, adapting them only to the specific dimensionalities. Please refer to \cref{app:baselines} for a description of where the respective nets appear.

\paragraph{Neural architectures.} 
We use the notation $A,B,C$, to indicate fully connected layers of $A, B, C$ neurons respectively. All models use exponentiated linear units~\citep{clevert2015} as activation functions between layers.
They follow the layer width and depth of \citet{shalit2017} as closely as possible.

\begin{center}
    \centering
    \begin{tabular}{lcl}
        \toprule
        Function & Mapping & Comments\\
        \midrule
         $\mu(\cdot,\cdot)$& $\dim(\mbx) + 1, 200,200,\dim(z),100,100,1$  \\
         $\mu_i(\cdot), i\in \{0,1\}$& $\dim(\mbx),200,200,\dim(z),100,100,1$ & \emph{TNet}\\
         $\mu_i(\cdot), i\in \{0,1\}$& $\dim(z),100,100,1$ & otherwise\\
         $h(\cdot)$ & $\dim(\mbx),200,200,\dim(z)$\\
         $h_i(\cdot), i\in \{1,\ldots, 5\}$ & $\dim(\mbx),200,200,s_i$ & $s_i$ defined in the hyperparameters \\
         $g(\cdot)$, $f_{z\to t}$ & $\dim(\mbz),1$ \\
         $\lambda(\cdot)$ & $\dim(\mbc),10,10,\dim(\mbz)$ \\
         $f_{z\to uz}(\cdot)$ & $\dim(\mbz),50$ \\
         $f_{u\to uz}(\cdot)$ & $\dim(\mbz),150$ \\
         $f_{uz\to x}(\cdot)$ & $200,200,\dim(\mbx)$ \\
         $f_{z\to x}(\cdot)$ & $\dim(\mbz),200,200,\dim(\mbx)$ \\
         $f_{z\to y}(\cdot)$ & $\dim(\mbz),100,100,1$ \\
         $\mu_t^u(\cdot),\sigma_t^u(\cdot)$ & $\dim(\mbx),200,200,\dim(\mbu)$ \\
         $\mu_t^z(\cdot),\sigma_t^z(\cdot)$ & $\dim(\mbx),200,200,\dim(\mbz)$ \\
         $\mu_t^{z_m}(\cdot),\sigma_t^{z_m}(\cdot)$ & $\dim(\mbm),200,200,\dim(\mbz^m)$ \\
         \bottomrule
    \end{tabular}
\end{center}

\paragraph{Hyperparameters and further settings.}
In the fully observed setup, the representation space $\mcZ$ is five-dimensional, with~$s = (1,1,2,1)$ to specify the split in the SNet representation space.  
The dimensionality of $\mcU$ is fixed at 50 throughout all experiments. 

Discrete covariates are treated as Bernoulli or categorical variables turned into one-hot encodings for all models in the case of categorical covariates. Continuous covariates are treated as normal variables with a homoscedastic noise model, with a variance parameter per covariate inferred via gradient descent. These are modeled as normal variables with a global precision parameter inferred via gradient descent.

We perform gradient descent via the Adam optimizer~\citep{kingma15} with a learning rate of $0.001$ and weight decay of $0.0001$. A random subset of 10\% of the training data in each of the replications is used as a validation set for early stopping. The upper limit for the maximal number of epochs was set to 500, which was never reached. 

For baselines that do not model missing values, we impute the empirical mean for missing continuous covariates and the median for missing binary ones. (Categorical covariates were assumed to always be observed for simplicity.)

\subsection{Semisynthetic Experimental Data}\label{sec:app-data}

Due to the fundamental problem of causality \citep{pearl2009}, the true treatment effect is not known for real data as the counterfactual outcome can never be observed. Any experimental evaluation is therefore constrained to rely on either completely synthetic or semi-synthetic data sets. In the main paper, we focus on the latter case and rely on real covariates with simulated treatment outcomes. This allows us to stay realistic with respect to covariate distributions, while still being able to generate factual and counterfactual outcomes to simulate various treatment effect scenarios. 

We rely on two sets of covariates for this task, which we describe below in greater detail. The first is based on covariates from the popular IHDP benchmark \citep{hill2011}. The second is a newly curated real-world data set combining RCT study data with electronic health records. We describe each of them in turn.

\subsubsection{IHDP}\label{sec:app-data-ihdp}

\paragraph{Data.} The Infant Health and Development Program (IHDP) data set as used by \citet{hill2011} consists of 25 covariates. 19 of them are binary, five of them continuous and one categorical.\footnote{The categorical covariate is modelled as continuous in most prior work.} This observational data set contains 139 treated children and 608 untreated ones after preprocessing, where \citet{hill2011} removes all children with nonwhite mothers from the treated group to create an artificial reduction in overlap between the two groups. 

\paragraph{Synthetic outcome creation.}
As is common, we follow \citet{hill2011}.
Given the covariates $\mbx$, an offset vector $\mbw$, whose entries all equal $0.5$, and a regression vector $\beta$, whose coefficients are randomly sampled from $(0,0.1,0.2,0.3,0.4)$ with probabilities $(0.6,0.1,0.1,0.1,0.1)$, the potential outcomes are generated as
\begin{equation*}
     Y(0) \sim\Norm(\exp((\mbx + \mbw)^\top \beta, 1)\quad\text{and}\quad Y(1) \sim \Norm(\mbx^\top \beta - \omega, 1),
\end{equation*}
where $\omega$ is an offset, chosen so that the true treatment effect on the treated is given as $4$. 
For the different scenarios described below, $\mbx$ is further modified before creating the synthetic outcomes $Y(0)$ and $Y(1)$.
We rely on the precomputed train/val/test splits provided by \citet{shalit2017}. The number of random data sets used in each experiment is marked in the respective result tables or figure captions.

We generate $\mbc$ by picking three random covariates from the 19 binary covariates for each seed. Note that given the randomness in which covariates are predictive, this does not guarantee that the chosen $\mbc$ provides good guidance to $\mbz$. Similar to how the researcher cannot know a priori which covariates are useful.

\paragraph{All vs Subset.}
The \emph{all} scenario relies on all 25 covariates $\mbx$ to create the synthetic outcomes. As the $\beta$ vector is randomly sampled and can contain zeros, this still leaves us with potentially fewer than 25 predictive covariates. 

Compared to this, \emph{subset} explicitly constraints to fewer than 25 covariates by choosing a random subset of $n_\text{pred}$ covariates for each seed before following the procedure described above. As before, $n_\text{pred}$ only provides us with an upper bound of predictive covariates. Throughout our experiments, we use $n_\text{pred} = 15$. 

\paragraph{High vs Low.} 
The \emph{high} setup refers to using the original $\mbx$ as defined by \citet{hill2011}. We keep the overlap as is, which has already been perturbed to a certain degree as discussed above. 

For \emph{low} we try to mimic real-world scenarios where the overlap tends to be even lower. 
To increase the divergence between the two data sources we modify the control group as follows. We shift the five continuous covariates as 
\begin{equation*}
    X_c \leftarrow X_c + s + \veps, \quad\text{where}\;\;\; s \sim \Ber(0.5)\cdot6 - 3, \quad \veps \sim \Norm(0, 9 ),
\end{equation*}
for the $c$-th continuous covariate. That is, $s \in \{-3,3\}$ with equal probability, and $\veps$ adds a large amount of variation to still keep some overlap.  Of the $19$ binary covariates, we pick a random subset of five increasing the probability of three of them to be one by $80\%$ and the probability of the other two to be $0$ by $80\%$.

\paragraph{Missingness.} 
We compare three settings of increasing missing not at random structure. 
In each of the three settings, the covariates for each observation are missing based on the following:
\begin{itemize}
    \item \emph{Strong}. For discrete covariates: If $x_i=1$ it has a probability of $p=0.2$ of being observed, if $x_1 = 0$ it has a probability of $p=0.3$ of being observed. A continuous covariate with $x_i>0$ has a probability of $p=0.1$ of being observed.
    \item \emph{Medium}. For discrete covariates: If $x_i=1$ it has a probability of $p=0.5$ of being observed, if $x_1 = 0$ it has a probability of $p=0.7$ of being observed. A continuous covariate with $x_i>0$ has a probability of $p=0.6$ of being observed.
    \item \emph{Weak}. For discrete covariates: If $x_i=1$ it has a probability of $p=0.9$ of being observed, if $x_1 = 0$ it has a probability of $p=0.9$ of being observed. A continuous covariate with $x_i>0$ has a probability of $p=0.9$ of being observed.
\end{itemize}
Missing discrete covariates are imputed by the median of the observed data set, missing continuous by zeros.

\paragraph{Survival analysis.}
Data generation details for the survival analysis experiments are given in \cref{app:survival}.

\subsubsection{Real-world data} 
\paragraph{Data.} 

This data set consists of two parts. The single-arm trial data is taken from \citet{Bakris2020}, a published RCT study, evaluating the treatment of patients suffering from chronic kidney disease and type 2 diabetes. For our purpose, only a subset of the treated group of patients is selected for the model to train on (N=833). 
The control group is constructed from electronic health records provided by Helsinki University Hospital, whose ethics committee gave ethical approval and a study permit. A set of 2646 patients was selected who roughly fulfilled the inclusion criteria of the RCT study. Although both data sets are originally longitudinal, the measurements closest to the index date are collected into a single set of covariates, leaving proper longitudinal modeling for future work. Note that while the index date is properly defined in the RCT it remains a noisy choice for the EHR data. For each patient, we pick the time a subset of the original inclusion criteria is fulfilled.
A subset of 100 binary covariates of the most prevalent diagnosis and medical history is selected (from approximately 4000), as well as $40$ continuous covariates (lab measurements age, weight, height, \ldots) giving us a total of 140 covariates. Due to their wide range of potential values, we standardize them by subtracting the median and scaling them by the interquartile range.\footnote{See \texttt{sklearn.preprocessing.RobustScaler} for an implementation.} 
As the EHR data contains a lot of missing covariates, we impute missing binary covariates via their median, while missing continuous ones are imputed with zeros (justified due to the prior standardization of the data). 
To create synthetic treatment outcomes, we require fully observed covariates, i.e., cannot model the true missingness pattern inherent in the model. 
We provide a longer discussion on the differences and similarities in the respective covariate distributions in \citet{Kurki2024}.

\paragraph{Synthetic outcome creation.}
We generate synthetic outcomes as follows. First, a random subset of $30$ covariates is selected from the original 140 and mapped with a single hidden layer neural network (30,10,$\dim(\mbz)$)\footnote{See \cref{sec:app-arch_hyper} for details on this notation.} into a five-dimensional representation space. The outcomes are then given as
\begin{equation*}
    Y(0) \sim \Norm\left(\mba^\top \mbv, 0.3^2\right)\qquad\text{and}\quad Y(1) \sim \Norm\left((\mba + 0.5)^\top \mbw - \omega, 0.3^2\right),
\end{equation*}
where $\mbv,\mbw \sim \Norm(0,\1)$, $\mba$ the five dimensional representation, and $\omega$ such that the ATT is four.

\subsection{Alternative Matching Approaches}\label{sec:app-metric}

\paragraph{Matching without replacement.}
Throughout all of our experiments, we conduct matching with replacement, i.e., we match control patients to treated patients independently of whether they have been matched before. While this ensures that every treated patient gets a control that is closest to them, it effectively reduces the number of used patients and might induce statistical problems due to multiple samples being identical. While this is the most common approach, some alternatives can be considered. The first one would be to pick patients greedily for a random permutation of the treated patients. While this is simple, it will depend on the order and thus won't be optimal. Another approach is to consider the matching as a task that aims to find the optimal minimal weight solution for a bipartite graph. Considering each group as one set of vertices with the edges between the groups given by the respective distances, the optimal matching is given by solving a straightforward linear program.\footnote{Our implementation relies on a SciPy~\citep{2020SciPy-NMeth} routine.}

It should be noted that these two approaches solve the task of reusing patients, but suffer from the problem that the chosen subset is now heavily interdependent, i.e., the choice of one patient is no longer independent of the other choices. We evaluate this latter approach on a subset of methods in \cref{tab:app-linearsum}.

\begin{table}[ht]
    \centering
    \caption{\emph{Alternative matching.} 
    We compare matching with replacement (\emph{with}) to matching without replacement (\emph{w/out}). Matching without replacement performs similar, or even improves upon the default matching with replacement approach.
    }\label{tab:app-linearsum}
    \adjustbox{max width=0.98\textwidth}{
    \begin{tabular}{lcccccccc}
    \toprule
    \emph{AE of ATT} & \multicolumn{4}{c}{all+high} & \multicolumn{4}{c}{subset+low} \\
    \cmidrule(lr){2-5}\cmidrule(lr){6-9}
    \textsc{Method} & \multicolumn{2}{c}{within sample} & \multicolumn{2}{c}{out-of-sample} & \multicolumn{2}{c}{within sample} & \multicolumn{2}{c}{out-of-sample} \\
    \cmidrule(lr){2-3}\cmidrule(lr){4-5}\cmidrule(lr){6-7}\cmidrule(lr){8-9}
    & with & w/out& with & w/out& with & w/out& with & w/out  \\
    \midrule
    VAE & ${0.085 \sd{\pm0.004}}$ & $\underline{0.083 \sd{\pm0.004}}$ &  ${0.266 \sd{\pm0.012}}$ &$\underline{0.255 \sd{\pm0.011}}$ & $\underline{0.264 \sd{\pm0.014}}$ &$0.282 \sd{\pm0.014}$ &$0.427 \sd{\pm0.022}$ & $\underline{0.368 \sd{\pm0.018}}$ \\
    CEVAE &  $\underline{0.052 \sd{\pm0.002}}$ &$\underline{0.052 \sd{\pm0.002}}$ &   ${0.174 \sd{\pm0.009}}$ &$\underline{0.173 \sd{\pm0.008}}$ &$\underline{0.150 \sd{\pm0.009}}$ & $0.192 \sd{\pm0.011}$ &$\underline{0.259 \sd{\pm0.015}}$&$0.263 \sd{\pm0.014}$\\
    TEDVAE &${0.052 \sd{\pm0.003}}$  & $\underline{0.050 \sd{\pm0.002}}$&   $\underline{0.179 \sd{\pm0.009}}$ &$0.180 \sd{\pm0.009}$ &$\underline{0.144 \sd{\pm0.009}}$ &$0.198 \sd{\pm0.010}$&$0.272 \sd{\pm0.016}$&$\underline{0.271 \sd{\pm0.014}}$\\
    Ours & $\underline{0.037 \sd{\pm0.002}}$ & $\underline{0.037 \sd{\pm0.002}}$&$0.113 \sd{\pm0.007}$&$\underline{0.106 \sd{\pm0.006}}$ &$0.114 \sd{\pm0.008}$ &$\underline{0.107 \sd{\pm0.008}}$&${0.190 \sd{\pm0.016}}$ &$\underline{0.180 \sd{\pm0.013}}$\\
    \quad+I & ${0.038 \sd{\pm0.002}}$&  $\underline{0.036 \sd{\pm0.002}}$ &${0.122 \sd{\pm0.007}}$&$\underline{0.116 \sd{\pm0.007}}$ & $0.110 \sd{\pm0.007}$ & $\underline{0.108 \sd{\pm0.008}}$ & ${0.176 \sd{\pm0.014}}$& $\underline{0.167 \sd{\pm0.011}}$\\
    \quad+sep &${0.037 \sd{\pm0.002}}$ & $\underline{0.035 \sd{\pm0.002}}$ &${0.111 \sd{\pm0.007}}$& $\underline{0.109 \sd{\pm0.007}}$ &${0.144 \sd{\pm0.009}}$ & $\underline{0.110 \sd{\pm0.008}}$ & $0.272 \sd{\pm0.016}$  & $\underline{0.174 \sd{\pm0.012}}$\\
    \bottomrule
    \end{tabular}}\\
    {\tiny{\emph{mean $\pm$ standard error over 300 random replications; the lower of the two means between each variation is \underline{underlined}}}}\\
\end{table}

\paragraph{Varying metrics.}
Within the main paper, we always rely on the Euclidian distance 
\begin{equation*}
    d(\mba,\mbb) = \sqrt{\sum_{i=1}^d (a_i - b_i)^2},
\end{equation*}
for $\mba,\mbb \in \mdR^d$.
However, given that our variational posteriors consist of distributions instead of point estimates, we can also rely on distributional distance metrics that take posterior variances into account. 

The \emph{Wasserstein distance} between two univariate Normal distributions is analytically tractable and given as 
\begin{equation*}
   W(\Norm(\mu_0,\sigma_0), \Norm(\mu_1,\sigma_1))^2 = (\mu_0 - \mu_1)^2 + (\sigma_0 - \sigma_1)^2. 
\end{equation*}

The squared \emph{Hellinger distance} between two univariate Normal distributions is given as 
\begin{equation*}
   H(\Norm(\mu_0,\sigma_0), \Norm(\mu_1,\sigma_1))^2 = 1 - \sqrt{\frac{2\sigma_0\sigma_1}{\sigma_0^2 + \sigma_1^2}}\exp\left(-\frac{(\mu_0 -\mu_1)^2}{4(\sigma_0^2 + \sigma_1^2)}\right).
\end{equation*}
Both Wasserstein and Hellinger factorize for mean-field normal distributions which is why we only state their univariate formulae.

For $q(\mbz_i) = \Norm(\mbz_i, \mu_i,\sigma_i^2)$, and we $q(\mbz_j) = \Norm(\mbz_j| \mu_j, \sigma_j^2)$, we 
compare the ATT performance for matching in the reduced setting via \emph{(i)} the Euclidean distance $d(\mu_i,\mu_j)$, \emph{(ii)} the Wasserstein distance $W(q(\mbz_i),q(\mbz_j))$, \emph{(iii)} the Hellinger distance $H(q(\mbz_i),q(\mbz_j))$, and \emph{(iv)} via the Euclidean distance on a propensity estimator $\hat \pi$, i.e., $d(\hat\pi(\mbz_i), \hat\pi(\mbz_j))$. 

As in the remainder of the paper, we stick to mean-field normal posteriors. 
However, the low-dimensional structure of $\mbz$ allows us to infer more complex posteriors without too many constraints due to computational or memory costs. E.g., by switching to a multivariate Normal, for which the two distances remain analytically tractable, while at the same time allowing for more structured variance estimation. We leave such extensions to future work. 

The results are summarized in \cref{app:matchmetric}.

\subsection{Predictive Latent Space $\mcZ$ Regularization}\label{sec:app-regularize}

As discussed in our main paper, one crucial assumption for causal inference from observational data is overlap, i.e., that $0<\pi(x)<1$, $\forall x$ where $\pi(x) = \mdP(T=1|X=x)$ is the propensity score. It is therefore common practice~\citep{shalit2017,johansson2016,lu2020} to further regularize the latent space, in our case $\mcZ$, on top of the existing regularization through the loss. In our case it is already implicitly regularized by the KL terms within our ELBO~\eqref{eq:elbo}, i.e., in the identifiable setup $\KL{q(\mbz)}{p(\mbz|\mbc)}$ and $\KL{q(\mbz)}{p(\mbz)}$ in the unidentifiable one. 
To further encourage overlap through additional regularization we can rely on several approaches. 
One could rely on distributional approximation methods, such as the maximum mean discrepancy method (MMD)~\citep{gretton12}, as was proposed, e.g., by \citet{johansson2016,shalit2017}.
MMD aims to minimize the distance between two aggregate variational posteriors
\begin{equation*}
    q_t(\mbz) \deq \frac{1}{N_t}\sum_{n:t_n=t} q(\mbz_n|\mbx_n), \qquad t \in \{0,1\},
\end{equation*}
by minimizing $\text{MMD}^2(q_0,q_1)$ defined as
\begin{equation*}
    \text{MMD}^2(q_0,q_1) = \Ep{q_0(z)q_0(z')}{k(z,z')}- 2\Ep{q_0(z)q_1(z')}{k(z,z')} + \Ep{q_1(z)q_1(z')}{k(z,z')},
\end{equation*}
where we use the exponentiated quadratic kernel as $k(\cdot,\cdot)$ and approximate the  expectations via sampling. 
The loss to be optimized is then given as 
\begin{equation*}
    \mcL_\text{elbo} + \kappa\text{MMD}^2(q_0,q_1), %
\end{equation*}
where $\kappa$ serves as a balancing factor. Throughout our experiments, we use a fixed $\kappa=1$

Another approach is to use critic-based approaches \citep[e.g.,][]{lu2020}, i.e., learning to fool a critic whose task is to differentiate between treated and control observations, as is commonly done in generative adversarial networks (GANs) \citep{goodfellow2014}.
Within this work, we rely primarily on the gradient reversal layer formulation introduced by~\citet{ganin2015} for GAN-based regularization.
An additional discriminator tries to classify instances $\mbz_n$ correctly as treated/control, i.e., to maximize the log-likelihood $\log p(t_n|\mbz_n)$. The main model tries to maximize its increasing overlap in the process. 
These two diverging objectives can be incorporated in our existing objective \eqref{eq:elbo} by relying on a pseudo-function 
\begin{equation*}
    R_\kappa(\mbz) \deq \mbz\quad\text{with}\quad \frac{dR_\kappa}{d\mbz} \deq -\kappa \1,
\end{equation*}
and using ${\tilde f(\mbz) \deq (f\circ R_\kappa)(\mbz)}$ for the likelihood ${p(t|\mbz) = \Ber(t|\sigma(\tilde f(\mbz))}$, and $f$ a neural network. 
The inherent objective of the discriminator $f$ is then to maximize $\log p(t|\mbz)$, while the objective of our generative model becomes a modification of the ELBO, i.e., its objective is to 
\begin{equation*}
        \text{maximize} \quad \mcL_\text{elbo} - 2\log p(t|\mbz),
\end{equation*}
where $\mcL_\text{elbo}$ is given by \eqref{eq:elbo}. We use $\kappa = 1$ in the experiments. 

These regularization approaches apply to the case where we want full overlap in the latent space $\mcZ$. This is the case for having access to outcome data from both groups, i.e., the single-arm patients as well as the control. For our estimators $\mu_t$ to be predictive of the counterfactual of the other group, we want the overlap between these groups to be maximal. 
However, if we only have access to outcome data $y$ for the control group, we only require the encodings of the treated group to overlap with the encodings of the control group, but not vice versa. I.e., as long as we can select a suitable subset of control patients, it is irrelevant how the remainder of the controls are encoded. It is therefore sometimes already sufficient to rely on the indirect regularization provided by the KL terms and the generative constraints to ensure a sufficient overlap.

We do this in the one-arm experiments for IHDP. Note, that this is also necessary to fulfill the assumptions of \citet{khemakhem2020}'s Theorem~4 (see above).

\subsection{Survival}\label{app:survival}

Our survival analysis experiment uses the IHDP covariates as its starting point after transforming them following the \emph{subset+low} strategy. We only experiment with right-censored observations.

Synthetic outcomes are then generated via the following steps, where we roughly follow \citet{Pölsterl_2019, manduchi2022a}.

\begin{enumerate}
    \item Sample $\beta$ as in the default IHDP setting from $(0,0.1,0.2,0.3,0.4)$ with probabilities $(0.6,0.1,0.1,0.1,0.1)$
    \item For every patient $n$ compute a risk score assuming they are treated $r_n^t$ or control $r_n^c$ as 
    \begin{align*}
        r_n^c &= \text{clip}\big(\text{soft}\big(\mbx_n^\top \beta + ((\mbx - 0.5)^\top \beta )^2\big)\big)\\
        r_n^t &= \text{clip}\big(\text{soft}\big(\mbx_n^\top \beta + |\mbx^\top \beta|\big)\big)
    \end{align*}
    where $\text{soft}(x) = \log(1 + \exp(x))$ is the softplus function, and we clip them to be within $[0.5,15]$. 
    \item For each $\mbr^c=(r_1^c,\ldots,r_N^c)$ and $\mbr^t=(r_1^t,\ldots,r_N^t)$, we, relying on a generic $\mbr$ for notational simplicity in the following list, create survival times $\mby$ and censoring indicators $\delta$ as follows
    \begin{enumerate}
        \item Define mean survival time $T_0$ ($=365$ in our experiments) and $p_\text{cens}$ the probability of being censored ($=0.3$ in our experiments).
        \item Let $\lambda_n = \exp(r_n)/T_0$, for $n=1,\ldots, N$.
        \item Let $u_n = -\frac{\log a_n}{\lambda_n}$, for $a_n \sim \Unif(0,1)$ for $n=1,\ldots,N$.
        \item Let $q_\text{cens}= q_{1 - p_\text{cens}}s(\mbu)$, where $\mbu = (u_1,\ldots, u_N)$ and $q_\alpha(\cdot)$ is the $\alpha$-th quantile.
        \item Sample $t_\text{cens} \sim \Unif(\min_n u_n, q_\text{cens})$.
        \item Let $\delta_n = \begin{cases}
            1, &\text{if $u_n \leq t_\text{cens}$}\\
            0, &\text{else}
        \end{cases}$, for $n \in \{1,\ldots, N\}$.
        \item Let $y_n = \begin{cases}
            u_n, &\text{if $\delta = 1$}\\
            t_\text{cens},&\text{else}
        \end{cases}$, for $n \in\{1,\ldots, N\}$.
    \end{enumerate}
    \item Normalize survival times $y$ to the interval $[0.001,1.001]$.
\end{enumerate}

During inference, each model has access to pre-treatment covariates for the treated group, i.e, $\mbx$, with post-treatment outcome data being hidden, and to pre-treatment covariates $\mbx$, as well as, (censored) survival times $y$ and event indicators $\delta$ for the control group. 

We model the likelihood $p(y|t,\mbz)$ for each model as a Weibull distribution, $\mcW(\lambda,k)$, whose density is given as 
\begin{equation*}
   f(y;\lambda,k) =  \frac k\lambda \left(\frac y\lambda\right)^{k-1}\exp\left(-\left(\frac y\lambda\right)^k\right)I(y\geq 0),
\end{equation*}
with an indicator function $I$. For a survival function $S(y|t,\mbz) = \int_y^\infty f(\tau;\lambda,k) \d\tau$, the likelihood is given as 
\begin{align*}
    p(y|t,\mbz) &= f(y; \lambda(\mbz),k)^\delta S(y|t,\mbz)^{1 - \delta}\\
    &= \left(\frac{k}{\lambda_t(\mbz)} \left(\frac{y}{\lambda_t(\mbz)}\right)^{k-1}\exp\left(-\left(\frac{y}{\lambda_t(\mbz)}\right)^k\right)\right)^\delta \left(\exp\left(-\left(\frac{y}{\lambda_t(\mbz)}\right)^k\right)\right)^{1 - \delta},
\end{align*}
where the shape parameter $k$ is a fixed hyperparameter ($k=1$ throughout our experiments), and ${\lambda_t(\mbz) = \log(1 + \exp(\mbz^\top \beta_t))}$. $\beta_0$ and $\beta_1$ are optimized via gradient descent together with the remaining parameters.

\subsection{Computation of Performance Metrics}

\paragraph{RMSE of CATE.} 
As specified in the main paper, the conditional average treatment effect is defined as
\begin{equation*}
    \E{Y(1) - Y(0)|X=x} = \mu_1(x) - \mu_0(x).
\end{equation*}
Given outcome information for treated ($t=1$) and untreated ($t=0$) observations, all our methods and baselines can infer estimators $\hat \mu_0(x)$ and $\hat \mu_1(x)$ to get CATE estimates at test time. Given that the outcomes are synthetically generated, we know the true CATE for each observation and can directly compute and report the RMSE. 

We report two RMSEs. First the \emph{within sample} RMSE. During training time, for observation $(\mbx_n,t_n,y_n)$, we only observe the factual outcome $y_n = Y(t_n)$. To estimate the CATE at test time, a model needs to correctly predict the unseen counterfactual $Y(1 - t_n)$ as well. The \emph{out-of-sample} RMSE refers to new covariates $\mbx_m$, i.e., the model has no never seen them, nor their factual outcome.

\paragraph{AE of ATT.}
The average treatment effect for the treated is defined as 
\begin{equation*}
    \E{Y(1) - Y(0)|T=1} = \E{Y(1)|T=1} - \E{Y(0)|T=1}.
\end{equation*}
In our specific setting, we assume that outcomes $y$ are only observed for the control group, i.e., for $t=0$. 

Once we are unblinded, we can estimate $\E{Y(1)|T=1}$ by the sample average $\frac{1}{N_1}\sum_{n=1:t_n=1}^N y_n$. 

Building an estimator for $\E{Y(0)|T=1}$ requires the selection of a suitable subset of control observations that match the characteristics of the treated observations. To avoid any statistical bias, we require this matching to be based only on pre-treatment information, i.e., only on observed covariates for the single-arm trial data. After matching, we estimate $\E{Y(0)|T=1}$ via the empirical average $\frac{1}{N_J}\sum_{j \in J}y_j$, where $J$ are the indices of the matched patients, and $N_J$ the number of matched samples.
After matching the treatment outcome is unblinded and the ATT can be estimated and compared with the synthetic true ATT.

\paragraph{Squared error for Time-to-event data.}
For our semi-synthetic survival experiment, the goal is to compare estimated hazard ratios. The evaluation proceeds in four steps: \emph{(i)} The true hazard ratio is estimated by fitting a Cox proportional hazards regression model on the factual and counterfactual survival curves of the single-arm group; \emph{(ii)} each model is fit using pre-treatment covariates for both groups and post-treatment survival outcomes for the control group; \emph{(iii)} a subset of control patients is selected and a second cox proportional hazards regression model is fit on the survival curve for the selected subset as well as the, now unblinded, observed factual outcomes of the treated group; \emph{(iv)} the squared difference between the two hazard ratio estimates is computed and reported. 

Cox models are fitted in R using the \texttt{survival} package by \citet{survival-package}.

\section{FURTHER EVALUATION}\label{sec:app-eval}
In this section we provide extended results on the experiments provided in the main paper.

\subsection{Runtime}\label{sec:app-runtime}
\paragraph{IHDP.} 
Given the small size of the models and the small size of the IHDP data set, models can be trained efficiently and fast on a modern CPU. Training a deterministic model takes about 30 seconds and about twice as much for a generative one. Differences between the individual deterministic/generative approaches are too minuscule to be relevant. These numbers apply to all three experimental setups, CATE estimation, ATT estimation, and survival analysis. 
Depending on the experimental setup we ran 100--300 replications.

\paragraph{Real-world data.}
Training on the real-world set takes about five minutes for a deterministic model and about twice as much for a generative one.

\subsection{Full Results For Both Arms}\label{sec:app-full}
We report results for all four settings on the IHDP data set in \cref{app-tap:resultfull}.
\begin{table*}[ht]
    \centering
    \caption{\emph{Outcome for both groups.} 
    Extended results related to \cref{tab:main-joint} in the main paper.
    }\label{app-tap:resultfull}
        \adjustbox{max width=0.98\textwidth}{
    \begin{tabular}{lcccccccccc}
    \toprule
    \emph{\small RMSE of CATE}&\multicolumn{2}{c}{all+high} &\multicolumn{2}{c}{all+reduced} &\multicolumn{2}{c}{subset+high} & \multicolumn{2}{c}{subset+reduced}&  \\
    \cmidrule(lr){2-3}\cmidrule(lr){4-5}\cmidrule(lr){6-7}\cmidrule(lr){8-9}
\textsc{Method} & within sample & out-of-sample & within sample& out-of-sample & within sample& out-of-sample & within sample& out-of-sample \\ 
    \midrule
    CFor & $0.556 \sd{\pm 0.000}$ & $0.5967 \sd{\pm 0.001}$  & $4.571 \sd{\pm 0.043}$ & $4.074 \sd{\pm 0.047}$ & $0.508 \sd{\pm 0.001}$ & $0.507 \sd{\pm 0.001}$ & $3.449\sd{\pm 0.052}$ & $2.597 \sd{\pm 0.031}$\\
    PScov & -- & -- & -- & -- & -- & -- & -- & --  \\
    PSpca & -- & -- & -- & -- & -- & -- & -- & -- \\
    PSlat & -- & -- & -- & -- & -- & -- & -- & -- \\
    SingleNet & $0.241 \sd{\pm 0.005}$ &$0.328 \sd{\pm 0.009}$  &$0.863 \sd{\pm 0.013}$ &$0.839 \sd{\pm 0.030}$ &$0.250 \sd{\pm 0.005} $&$0.274 \sd{\pm 0.006} $&$0.924 \sd{\pm 0.016}$ &$0.878 \sd{\pm 0.027}$\\
    TNet & $0.175 \sd{\pm 0.003}$ &$0.275 \sd{\pm 0.008}$  &$0.210 \sd{\pm 0.008}$ &$0.346 \sd{\pm 0.020}$ &$0.217 \sd{\pm 0.004} $&$0.246 \sd{\pm 0.005} $&$0.278 \sd{\pm 0.009}$ &$0.328 \sd{\pm 0.017}$\\
    TARNet& $0.177 \sd{\pm 0.004}$ &$0.280 \sd{\pm 0.008}$  &$0.220 \sd{\pm 0.009}$&$0.343 \sd{\pm 0.019}$ &$0.221 \sd{\pm 0.004} $&$0.251 \sd{\pm 0.005} $&$0.272 \sd{\pm 0.009}$ &$0.333 \sd{\pm 0.019}$\\
    CFRNet& $0.171 \sd{\pm 0.003}$ &$0.279 \sd{\pm 0.008}$  &$0.297 \sd{\pm 0.009}$&$0.397 \sd{\pm 0.020}$ &$0.179 \sd{\pm 0.003} $&$0.210 \sd{\pm 0.005} $&$0.279 \sd{\pm 0.008}$ &$0.338 \sd{\pm 0.017}$\\
    SNet&$0.168 \sd{\pm 0.003}$ &$0.264 \sd{\pm 0.008}$  &$0.228 \sd{\pm 0.009} $&$\mathbf{0.299 \sd{\pm 0.018}}$ &$0.159 \sd{\pm 0.003} $&$0.194 \sd{\pm 0.005} $&$0.211 \sd{\pm 0.008}$ &$\underline{\mathbf{0.243 \sd{\pm 0.016}}}$\\
    VAE & -- & -- & -- & -- & -- & -- & -- & -- \\
    CEVAE& $0.182 \sd{\pm 0.002}$ &$0.287 \sd{\pm 0.007}$ &$0.224 \sd{\pm 0.008} $&$0.348 \sd{\pm 0.019} $&$0.223 \sd{\pm 0.003} $&$0.256 \sd{\pm 0.004} $&$0.288 \sd{\pm 0.009}$ &$0.350 \sd{\pm 0.017}$\\
    TEDVAE& $0.176 \sd{\pm 0.002}$ &$0.283 \sd{\pm 0.007}$& $0.246 \sd{\pm 0.008} $&$0.344 \sd{\pm 0.019} $&$0.202 \sd{\pm 0.002} $&$0.238 \sd{\pm 0.004} $&$0.293 \sd{\pm 0.009}$ &$0.336 \sd{\pm 0.018}$\\ \midrule
    Ours& $0.152 \sd{\pm 0.002}$ &$0.267 \sd{\pm 0.008}$  &$0.174 \sd{\pm 0.006} $&$0.304 \sd{\pm 0.019} $&$0.161 \sd{\pm 0.003} $&$0.202 \sd{\pm 0.005} $&$0.201 \sd{\pm 0.007}$ &$0.272 \sd{\pm 0.016}$\\
    \quad+I& $0.141 \sd{\pm 0.002}$ &${0.251 \sd{\pm 0.008}}$  &$\underline{\mathbf{0.159 \sd{\pm 0.005}}} $&$\mathbf{0.293 \sd{\pm 0.017}} $&$0.157 \sd{\pm 0.003} $&$0.195 \sd{\pm 0.005} $&$0.194 \sd{\pm 0.007}$ &$0.262 \sd{\pm 0.016}$\\
    \quad+sep& $0.151 \sd{\pm 0.002}$ &$0.267 \sd{\pm 0.008}$  &$0.184 \sd{\pm 0.007} $&$0.310 \sd{\pm 0.019} $&$0.162 \sd{\pm 0.002} $&$0.204 \sd{\pm 0.005} $&$0.209 \sd{\pm 0.007}$ &$0.278 \sd{\pm 0.016}$\\
    \quad+sep+I& $0.143 \sd{\pm 0.002}$ &$0.255 \sd{\pm 0.007}$ &$\mathbf{0.167 \sd{\pm 0.006}} $&$\mathbf{0.294 \sd{\pm 0.017}} $&$0.162 \sd{\pm 0.003} $&${0.202 \sd{\pm 0.005}} $&$0.197 \sd{\pm 0.007}$ &$0.266 \sd{\pm 0.017}$\\
    \quad+snet& $0.141 \sd{\pm 0.002}$ &$0.256 \sd{\pm 0.008}$ &$0.183 \sd{\pm 0.007} $&$0.308 \sd{\pm 0.018} $&$0.145 \sd{\pm 0.002} $&$\mathbf{0.184 \sd{\pm 0.005}} $&$\mathbf{0.185 \sd{\pm 0.007}}$ &$\mathbf{0.253 \sd{\pm 0.017}}$\\
    \quad+snet+I& $\underline{\mathbf{0.130 \sd{\pm 0.002}}}$ &$\underline{\mathbf{0.242 \sd{\pm 0.008}}}$ &$\mathbf{0.161 \sd{\pm 0.005}} $&$\underline{\mathbf{0.288 \sd{\pm 0.017}}} $&$\underline{\mathbf{0.139 \sd{\pm 0.002}}} $&$\mathbf{0.182 \sd{\pm 0.005}} $&$\mathbf{0.181 \sd{\pm 0.006}}$ &$\mathbf{0.250 \sd{\pm 0.017}}$\\
    \quad+snet+sep& $0.143 \sd{\pm 0.002}$ &$0.254 \sd{\pm 0.008}$  &$0.168 \sd{\pm 0.006} $&$0.297 \sd{\pm 0.018} $&$\mathbf{0.143 \sd{\pm 0.002}} $&$\mathbf{0.186 \sd{\pm 0.005}} $&$0.191 \sd{\pm 0.007}$ &$0.258 \sd{\pm 0.017}$\\
    \quad+snet+sep+I& ${\mathbf{0.131 \sd{\pm 0.002}}}$ &${\mathbf{0.247 \sd{\pm 0.008}}}$ &$\mathbf{0.160 \sd{\pm 0.005}} $&$\mathbf{0.290 \sd{\pm 0.017}} $&$\mathbf{0.142 \sd{\pm 0.002}} $&$\underline{\mathbf{0.182 \sd{\pm 0.005}}} $&$\underline{\mathbf{0.180 \sd{\pm 0.007}}}$ &${\mathbf{0.251 \sd{\pm 0.017}}}$\\
    \quad+tedvae& $0.143 \sd{\pm 0.002}$ &$0.257 \sd{\pm 0.008}$ &$0.186 \sd{\pm 0.007} $&$0.310 \sd{\pm 0.018} $&$0.146 \sd{\pm 0.002} $&$\mathbf{0.184 \sd{\pm 0.005}} $&$\mathbf{0.187 \sd{\pm 0.007}}$ &$0.258 \sd{\pm 0.019}$\\
    \quad+tedvae+I& $0.142 \sd{\pm 0.002}$ &$0.255 \sd{\pm 0.008}$ &$0.174 \sd{\pm 0.007} $&$0.305 \sd{\pm 0.019} $&$\mathbf{0.142 \sd{\pm 0.002}} $&$\mathbf{0.182 \sd{\pm 0.005}} $&$0.191 \sd{\pm 0.006}$ &$\mathbf{0.253 \sd{\pm 0.016}}$\\
    \quad+tedvae+sep& $0.143 \sd{\pm 0.002}$ & $0.257 \sd{\pm 0.008}$&$0.192 \sd{\pm 0.007} $&$0.317 \sd{\pm 0.019} $&$0.146 \sd{\pm 0.002} $&$0.188 \sd{\pm 0.005} $& $0.200 \sd{\pm 0.007}$ & $0.263\sd{\pm0.017}$\\
    \quad+tedvae+sep+I& $0.138\sd{\pm0.002}$ & $0.250\sd{\pm 0.008}$&$0.181 \sd{\pm 0.006} $&$0.308 \sd{\pm 0.018} $&$\mathbf{0.142 \sd{\pm 0.002}} $&$\mathbf{0.185 \sd{\pm 0.005}} $& $0.190\sd{\pm 0.007}$ & $0.261 \sd{\pm 0.016}$\\
    \bottomrule
    \end{tabular}}\\
    \tiny{\emph{mean $\pm$ standard error over 300 random replications; statistically significant best models marked \textbf{bold}; lowest mean \underline{underlined}}}
\end{table*}

\subsection{Full Results For Single-arms}\label{sec:app-partial}
We report results for all four settings on the IHDP data set in \cref{app-tap:resultsingle}.
\begin{table*}[ht]
    \centering
    \caption{\emph{Outcome only for the control group.} 
    Extended results related to \cref{tab:main-joint} in the main paper.
    }\label{app-tap:resultsingle}
        \adjustbox{max width=0.98\textwidth}{
    \begin{tabular}{lcccccccccc}
    \toprule
    \emph{\small AE of ATT}&\multicolumn{2}{c}{all+high} &\multicolumn{2}{c}{all+reduced} &\multicolumn{2}{c}{subset+high} & \multicolumn{2}{c}{subset+reduced}&  \\
    \cmidrule(lr){2-3}\cmidrule(lr){4-5}\cmidrule(lr){6-7}\cmidrule(lr){8-9}
\textsc{Method} & within sample & out-of-sample & within sample& out-of-sample & within sample& out-of-sample & within sample& out-of-sample \\ 
    \midrule
    CFor & -- & -- & -- & -- & -- & -- & -- & --\\
    PScov & $0.115 \sd{\pm0.005}$ &$0.336 \sd{\pm0.017}$ & $0.648 \sd{\pm0.082}$ & $0.804 \sd{\pm0.265}$ &$0.111 \sd{\pm0.005}$ & $0.373 \sd{\pm0.019}$ &$0.691 \sd{\pm0.059}$ &$0.702 \sd{\pm0.063}$ \\
    PSpca &  $0.096 \sd{\pm0.005}$ &$0.300 \sd{\pm0.014}$ & $0.556 \sd{\pm0.056}$ & $0.811 \sd{\pm0.265}$ &$0.115 \sd{\pm0.005}$&$0.310 \sd{\pm0.014}$&$0.496 \sd{\pm0.036}$ &$0.613 \sd{\pm0.051}$\\
    PSlat & $0.084 \sd{\pm0.004}$ &$0.187 \sd{\pm0.013}$ & $0.501 \sd{\pm0.039}$ & $0.290 \sd{\pm0.024}$ & $0.096 \sd{\pm0.006}$& $0.260 \sd{\pm0.016}$ &$0.276 \sd{\pm0.025}$ &$0.272 \sd{\pm0.023}$ \\
    SingleNet &-- & -- & -- & -- & -- & -- & -- & -- \\
    TNet &-- & -- & -- & -- & -- & -- & -- & -- \\
    TARNet& $0.043 \sd{\pm0.003}$ &$0.130 \sd{\pm0.008}$ & $0.087 \sd{\pm0.008}$ &$0.108 \sd{\pm0.007}$ &$\mathbf{0.075 \sd{\pm0.004}}$ &$0.223 \sd{\pm0.012}$ &$0.140 \sd{\pm0.011}$ &$0.239 \sd{\pm0.016}$\\
    CFRNet&$0.042 \sd{\pm0.002}$ &$0.131 \sd{\pm0.007}$ & $0.321 \sd{\pm0.015}$ &$0.327 \sd{\pm0.016}$ &$0.080 \sd{\pm0.004}$ &$0.224 \sd{\pm0.012}$ &$0.345 \sd{\pm0.016}$ &$0.370 \sd{\pm0.017}$\\
    SNet& $0.043 \sd{\pm0.003}$ &$0.137 \sd{\pm0.007}$&$0.111 \sd{\pm0.010}$ &$0.178 \sd{\pm0.013}$ &$\mathbf{0.075 \sd{\pm0.004}}$ & $0.224 \sd{\pm0.012}$ & $0.221 \sd{\pm0.015}$ & $0.366 \sd{\pm0.026}$\\
    VAE & $0.085 \sd{\pm0.004}$ &$0.266 \sd{\pm0.012}$ &$0.280 \sd{\pm0.015}$ &$0.360 \sd{\pm0.028}$ &$0.094 \sd{\pm0.004}$ &$0.271 \sd{\pm0.014}$ &$0.264 \sd{\pm0.014}$ &$0.427 \sd{\pm0.022}$ \\
    CEVAE& $0.052 \sd{\pm0.002}$ &$0.174 \sd{\pm0.009}$  &$0.096 \sd{\pm0.006}$ &$0.178 \sd{\pm0.011}$ &$0.086 \sd{\pm0.004}$ & $0.234 \sd{\pm0.012}$ & $0.150 \sd{\pm0.009}$ &$0.259 \sd{\pm0.015}$ \\
    TEDVAE& $0.052 \sd{\pm0.003}$ &$0.179 \sd{\pm0.009}$ &$0.108 \sd{\pm0.007}$ &$0.188 \sd{\pm0.012}$ &$0.082 \sd{\pm0.004}$ &$0.233 \sd{\pm0.012}$ &$0.144 \sd{\pm0.009}$ & $0.272 \sd{\pm0.016}$\\
    \midrule
    Ours& $\mathbf{0.037 \sd{\pm0.002}}$ & $\mathbf{0.113 \sd{\pm0.007}}$  &$0.074 \sd{\pm0.006}$ &$\mathbf{0.091 \sd{\pm0.008}}$ & $\mathbf{0.074 \sd{\pm0.004}}$ &$\mathbf{0.212 \sd{\pm0.011}}$ &$\mathbf{0.114 \sd{\pm0.008}}$ &$\mathbf{0.190 \sd{\pm0.016}}$ \\
    \quad+I& $\mathbf{0.038 \sd{\pm0.002}}$ &$\mathbf{0.122 \sd{\pm0.007}}$  &$\mathbf{0.055 \sd{\pm0.005}}$ &$\mathbf{0.080 \sd{\pm0.006}}$ &$0.079 \sd{\pm0.004}$ &$0.229 \sd{\pm0.011}$ &$\mathbf{0.110 \sd{\pm0.007}}$ &$\underline{\mathbf{0.176 \sd{\pm0.014}}}$\\
    \quad+sep& $\mathbf{0.037 \sd{\pm0.002}}$ &$\mathbf{0.111 \sd{\pm0.007}}$ &$\mathbf{0.063 \sd{\pm0.005}}$ &$\mathbf{0.083 \sd{\pm0.006}}$ &$\mathbf{0.074 \sd{\pm0.004}}$ &$0.222 \sd{\pm0.012}$ &${0.117 \sd{\pm0.009}}$ &$\mathbf{0.193 \sd{\pm0.016}}$\\
    \quad+sep+I& $\underline{\mathbf{0.035 \sd{\pm0.002}}}$ &$\mathbf{0.117 \sd{\pm0.008}}$ &$\mathbf{0.055 \sd{\pm0.005}}$ &$\underline{\mathbf{0.078 \sd{\pm0.006}}}$ &$\mathbf{0.074 \sd{\pm0.004}}$ &$0.226 \sd{\pm0.012}$ &${\mathbf{0.104 \sd{\pm0.007}}}$ &${\mathbf{0.185 \sd{\pm0.015}}}$\\
    \quad+snet&$\mathbf{0.038 \sd{\pm0.002}}$ &$\mathbf{0.117 \sd{\pm0.007}}$ & $0.091 \sd{\pm0.008}$ &$0.169 \sd{\pm0.012}$ &$\mathbf{0.076 \sd{\pm0.004}}$ &$\mathbf{0.217 \sd{\pm0.013}}$ &$0.194 \sd{\pm0.014}$ &$0.322 \sd{\pm0.020}$\\
    \quad+snet+I& $\mathbf{0.038 \sd{\pm0.002}}$ &$\mathbf{0.120 \sd{\pm0.008}}$ &$0.094 \sd{\pm0.007}$ &$0.159 \sd{\pm0.012}$ &$\mathbf{0.077 \sd{\pm0.004}}$ &$\mathbf{0.212 \sd{\pm0.011}}$ &$0.190 \sd{\pm0.012}$ &$0.322 \sd{\pm0.023}$\\
    \quad+snet+sep& $0.040 \sd{\pm0.003}$ &$\underline{\mathbf{0.109 \sd{\pm0.007}}}$  &$0.095 \sd{\pm0.007}$ &$0.173 \sd{\pm0.012}$ &$\underline{\mathbf{0.071 \sd{\pm0.004}}}$ & $\mathbf{0.211 \sd{\pm0.011}}$ &$0.179 \sd{\pm0.013}$ &$0.316 \sd{\pm0.020}$ \\
    \quad+snet+sep+I& $0.039 \sd{\pm0.003}$ &$\mathbf{0.115 \sd{\pm0.007}}$  &  $0.085 \sd{\pm0.006}$ &$0.156 \sd{\pm0.011}$ &$\mathbf{0.076 \sd{\pm0.004}}$  & $\mathbf{0.211 \sd{\pm0.011}}$ & $0.196 \sd{\pm0.015}$ &$0.292 \sd{\pm0.020}$\\
    \quad+tedvae& $\mathbf{0.039 \sd{\pm0.003}}$ &$\mathbf{0.119 \sd{\pm0.007}}$ &$0.076 \sd{\pm0.006}$ & $0.094 \sd{\pm0.007}$ &$\mathbf{0.074 \sd{\pm0.004}}$ &$\underline{\mathbf{0.196 \sd{\pm0.011}}}$ &$\mathbf{0.103 \sd{\pm0.008}}$ &$\mathbf{0.189 \sd{\pm0.015}}$\\
    \quad+tedvae+I& $\mathbf{0.036 \sd{\pm0.002}}$ & $0.124 \sd{\pm0.007}$ &$\underline{\mathbf{0.054 \sd{\pm0.004}}}$ & $\mathbf{0.085 \sd{\pm0.006}}$ & $\mathbf{0.072 \sd{\pm0.004}}$ &$0.234 \sd{\pm0.013}$ & $\underline{\mathbf{0.102 \sd{\pm0.007}}}$ &$\mathbf{0.191 \sd{\pm0.017}}$ \\
    \quad+tedvae+sep& $0.039 \sd{\pm0.002}$ &$0.125 \sd{\pm0.007}$ &$0.067 \sd{\pm0.005}$ &$\mathbf{0.089 \sd{\pm0.007}}$ & $\mathbf{0.075 \sd{\pm0.004}}$ &$\mathbf{0.209 \sd{\pm0.011}}$ &$\mathbf{0.112 \sd{\pm0.009}}$ &$0.197 \sd{\pm0.014}$ \\
    \quad+tedvae+sep+I& $0.039 \sd{\pm0.002}$ &$0.124 \sd{\pm0.007}$ &$\mathbf{0.061 \sd{\pm0.004}}$ &$\mathbf{0.091 \sd{\pm0.007}}$ &$\mathbf{0.074 \sd{\pm0.004}}$ &$0.218 \sd{\pm0.012}$ &$\mathbf{0.104 \sd{\pm0.008}}$  &$\mathbf{0.173 \sd{\pm0.012}}$\\
    \bottomrule
    \end{tabular}}\\
    \tiny{\emph{mean $\pm$ standard error over 300 random replications; statistically significant best models marked \textbf{bold}; lowest mean \underline{underlined}}}
\end{table*}

\subsection{Missingness}\label{sec:app-missingness}
We report results for various degrees of missingness in \cref{app-tab:missing}.

\begin{table*}[ht]
    \centering
    \caption{\emph{Missingness.} 
    Increasing the amount of missing covariates decreases performance as expected. Modeling this in our proposed approach recovers this to a certain degree.
    }
    \label{app-tab:missing}
    \vspace{0.5em}
    \textit{(a) outcome for both groups (all + high)\\}
    \adjustbox{max width=0.99\textwidth}{
    \begin{tabular}{lcccccccccccc}
    \toprule
    \emph{\small RMSE of CATE}&\multicolumn{2}{c}{none}&    \multicolumn{2}{c}{weak} & \multicolumn{2}{c}{medium} & \multicolumn{2}{c}{strong} \\
    \cmidrule(lr){2-3}\cmidrule(lr){4-5}\cmidrule(lr){6-7}\cmidrule(lr){8-9}
\textsc{Method} & within sample & out-of-sample & within sample& out-of-sample& within sample& out-of-sample & within sample& out-of-sample\\
    \midrule
         TARNet&$0.176 \sd{\pm 0.006} $&$0.263 \sd{\pm 0.012} $&$0.304 \sd{\pm 0.006} $&$0.314 \sd{\pm 0.010} $&$0.462 \sd{\pm 0.011} $&$0.463 \sd{\pm 0.019} $&$0.531 \sd{\pm 0.014} $&$0.607 \sd{\pm 0.025} $\\
         CFRNet&$0.170 \sd{\pm 0.005} $&$0.258 \sd{\pm 0.012} $&$0.302 \sd{\pm 0.007} $&$0.307 \sd{\pm 0.011} $&$0.457 \sd{\pm 0.011} $&$0.452 \sd{\pm 0.019} $&$0.527 \sd{\pm 0.014} $&$0.600 \sd{\pm 0.025} $\\
         SNet&$0.172 \sd{\pm 0.005} $&$0.254 \sd{\pm 0.013} $&$0.301 \sd{\pm 0.007} $&$0.293 \sd{\pm 0.012} $&$0.465 \sd{\pm 0.012} $&$0.444 \sd{\pm 0.018} $&$0.536 \sd{\pm 0.015} $&$0.644 \sd{\pm 0.027} $\\
         CEVAE&$0.185 \sd{\pm 0.005} $&$0.277 \sd{\pm 0.012} $&$0.295 \sd{\pm 0.005} $&$0.322 \sd{\pm 0.010} $&$0.438 \sd{\pm 0.009} $&$0.482 \sd{\pm 0.017} $&$0.511 \sd{\pm 0.012} $&$0.635 \sd{\pm 0.022} $\\
         TEDVAE&$0.176 \sd{\pm 0.004} $&$0.270 \sd{\pm 0.012} $&$0.297 \sd{\pm 0.005} $&$0.318 \sd{\pm 0.010} $&$0.448 \sd{\pm 0.011} $&$0.464 \sd{\pm 0.018} $&$0.514 \sd{\pm 0.013} $&$0.629 \sd{\pm 0.023} $\\
         Our &$0.152 \sd{\pm 0.004} $&$0.249 \sd{\pm 0.012} $&$0.297 \sd{\pm 0.005} $&$0.318 \sd{\pm 0.010} $&$0.412 \sd{\pm 0.010} $&$0.464 \sd{\pm 0.019} $&$0.492 \sd{\pm 0.013} $&$0.632 \sd{\pm 0.026} $\\
         \quad+I&$0.144 \sd{\pm 0.004} $&$0.236 \sd{\pm 0.013} $&$0.268 \sd{\pm 0.006} $&$0.301 \sd{\pm 0.010} $&$0.412 \sd{\pm 0.011} $&$0.453 \sd{\pm 0.017} $&$0.489 \sd{\pm 0.013} $&$0.621 \sd{\pm 0.026} $\\
         \quad+sep&$0.149 \sd{\pm 0.004} $&$0.245 \sd{\pm 0.013} $&$0.267 \sd{\pm 0.005} $&$0.298 \sd{\pm 0.011} $&$0.412 \sd{\pm 0.011} $&$0.459 \sd{\pm 0.019} $&$0.490 \sd{\pm 0.013} $&$0.622 \sd{\pm 0.025} $\\
         \quad+I+sep&$0.144 \sd{\pm 0.004} $&$0.243 \sd{\pm 0.013} $&$0.268 \sd{\pm 0.006} $&$0.300 \sd{\pm 0.011} $&$0.411 \sd{\pm 0.011} $&$0.452 \sd{\pm 0.018} $&$0.488 \sd{\pm 0.013} $&$0.629 \sd{\pm 0.026} $\\
         \midrule
         Our+mask & -- & -- &$0.224 \sd{\pm 0.004} $&$0.275 \sd{\pm 0.012} $&$0.331 \sd{\pm 0.006} $&$0.348 \sd{\pm 0.013} $& $0.430 \sd{\pm 0.010} $&$0.518 \sd{\pm 0.019} $\\
         \quad+I& -- & -- &$0.221 \sd{\pm 0.004} $&$0.264 \sd{\pm 0.011} $&$0.327 \sd{\pm 0.007} $&$0.356 \sd{\pm 0.014} $&$0.415 \sd{\pm 0.010} $&$0.506 \sd{\pm 0.021} $\\
         \quad+sep& -- & -- &$0.223 \sd{\pm 0.005} $&$0.267 \sd{\pm 0.011} $&$0.329 \sd{\pm 0.006} $&$0.348 \sd{\pm 0.014} $&$0.428 \sd{\pm 0.010} $&$0.495 \sd{\pm 0.017} $\\
         \quad+I+sep& -- & -- &$0.217 \sd{\pm 0.004} $&$0.257 \sd{\pm 0.011} $&$0.326 \sd{\pm 0.007} $&$0.345 \sd{\pm 0.013} $&$0.426 \sd{\pm 0.010} $&$0.492 \sd{\pm 0.018} $\\
         \bottomrule
    \end{tabular}}\\
    {\tiny{\emph{mean $\pm$ standard error over 100 random replications}}}\\
    \vspace{0.5em}
    \textit{(b) outcome only for the control group (subset + low)}\\
    \adjustbox{max width=0.99\textwidth}{
    \begin{tabular}{lcccccccccccc}
    \toprule
    \emph{\small AE of ATT}&\multicolumn{2}{c}{none}&    \multicolumn{2}{c}{weak} & \multicolumn{2}{c}{medium} & \multicolumn{2}{c}{strong} \\
    \cmidrule(lr){2-3}\cmidrule(lr){4-5}\cmidrule(lr){6-7}\cmidrule(lr){8-9}
\textsc{Method} & within sample & out-of-sample & within sample& out-of-sample& within sample& out-of-sample & within sample& out-of-sample\\
    \midrule
         TARNet&$0.138 \sd{\pm0.016}$ &$0.172 \sd{\pm0.018}$ & $0.216 \sd{\pm0.016}$ &$0.238 \sd{\pm0.024}$ &$0.289 \sd{\pm0.019}$ &$0.196 \sd{\pm0.020}$ &$0.448 \sd{\pm0.030}$ &$0.266 \sd{\pm0.027}$\\
         CFRNet&$0.415 \sd{\pm0.031}$ &$0.414 \sd{\pm0.036}$ &$0.405 \sd{\pm0.029}$ &$0.428 \sd{\pm0.031}$ &$0.318 \sd{\pm0.023}$ &$0.335 \sd{\pm0.029}$ &$0.413 \sd{\pm0.027}$ &$0.313 \sd{\pm0.030}$\\
         SNet&$0.177 \sd{\pm0.020}$ &$0.240 \sd{\pm0.024}$ & $0.248 \sd{\pm0.021}$ &$0.294 \sd{\pm0.027}$ & $0.305 \sd{\pm0.021}$ &$0.291 \sd{\pm0.026}$ & $0.434 \sd{\pm0.026}$ &$0.390 \sd{\pm0.035}$  \\
         CEVAE&$0.241 \sd{\pm0.021}$ &$0.274 \sd{\pm0.024}$ &$0.268 \sd{\pm0.021}$ &$0.353 \sd{\pm0.028}$ &$0.289 \sd{\pm0.021}$ &$0.302 \sd{\pm0.025}$ &$0.397 \sd{\pm0.028}$ &$0.312 \sd{\pm0.033}$\\
         TEDVAE&$0.238 \sd{\pm0.020}$ &$0.286 \sd{\pm0.025}$ &$0.267 \sd{\pm0.020}$ &$0.331 \sd{\pm0.027}$ &$0.289 \sd{\pm0.022}$ &$0.274 \sd{\pm0.023}$ &$0.409 \sd{\pm0.027}$ &$0.303 \sd{\pm0.034}$\\
         Our &$0.106 \sd{\pm0.015}$ &$0.160 \sd{\pm0.018}$ &$0.192 \sd{\pm0.014}$ &$0.215 \sd{\pm0.024}$ &$0.271 \sd{\pm0.021}$ &$0.246 \sd{\pm0.020}$ &$0.442 \sd{\pm0.028}$ &$0.291 \sd{\pm0.027}$\\
         \quad+I&$0.109 \sd{\pm0.016}$ &$0.185 \sd{\pm0.020}$ &$0.174 \sd{\pm0.013}$ &$0.214 \sd{\pm0.020}$ & $0.280 \sd{\pm0.021}$ &$0.236 \sd{\pm0.021}$ &$0.418 \sd{\pm0.027}$ &$0.365 \sd{\pm0.059}$\\
         \quad+sep& $0.104 \sd{\pm0.016}$ & $0.181 \sd{\pm0.019}$ &$0.193 \sd{\pm0.015}$ &$0.229 \sd{\pm0.023}$ &$0.288 \sd{\pm0.021}$ &$0.245 \sd{\pm0.021}$ &$0.445 \sd{\pm0.029}$ & $0.268 \sd{\pm0.026}$\\
         \quad+I+sep&$0.115 \sd{\pm0.017}$ &$0.162 \sd{\pm0.019}$ &$0.191 \sd{\pm0.015}$ &$0.331 \sd{\pm0.027}$ &$0.260 \sd{\pm0.020}$ &$0.215 \sd{\pm0.021}$ &$0.433 \sd{\pm0.030}$ &$0.332 \sd{\pm0.056}$ \\
         \midrule
         Our+mask& -- & -- &$0.150 \sd{\pm0.013}$ &$0.200 \sd{\pm0.021}$ &$0.197 \sd{\pm0.014}$ &$0.171 \sd{\pm0.019}$ &$0.378 \sd{\pm0.023}$ &$0.218 \sd{\pm0.022}$\\
         \quad+I& -- & -- &$0.137 \sd{\pm0.014}$ &$0.171 \sd{\pm0.020}$ &$0.176 \sd{\pm0.014}$ &$0.178 \sd{\pm0.018}$ &$0.337 \sd{\pm0.023}$ &$0.239 \sd{\pm0.025}$\\
         \quad+sep& -- & -- &$0.143 \sd{\pm0.012}$ &$0.176 \sd{\pm0.019}$ & $0.203 \sd{\pm0.015}$ &$0.165 \sd{\pm0.018}$ &$0.375 \sd{\pm0.024}$ &$0.230 \sd{\pm0.022}$\\
         \quad+I+sep& -- & -- &$0.138 \sd{\pm0.013}$ &$0.175 \sd{\pm0.019}$ &$0.179 \sd{\pm0.014}$ &$0.186 \sd{\pm0.019}$ &$0.337 \sd{\pm0.021}$ & $0.228 \sd{\pm0.027}$\\
         \bottomrule
    \end{tabular}}\\
    {\tiny{\emph{mean $\pm$ standard error over 100 random replications}}}\\
\end{table*}

\subsection{Matching Metrics}\label{app:matchmetric}

We evaluate two different scenarios on a subset of the methods discussed.  ATT estimation in \emph{(i)} the \emph{all+high} scenario, and \emph{(ii)} in the \emph{subset+low} scenario.
Adding distributional information within the mean-field assumption of $q(\mbZ)$ provides little benefit and tends to even hurt performance. Matching via a propensity score estimator in the latent space is never competitive in our setting.
The results following the current mean-field assumption are summarized in \cref{tab:app-tabmetric}. 

\begin{table}[ht]
    \centering
    \caption{\emph{Varying metrics for matching.} 
    We compare relying on the Euclidean metric using the posterior mean, to using distributional metrics or propensity score (PS) matching in the latent space.  
    }\label{tab:app-tabmetric}
    \emph{(i) ATT estimation for \emph{all + high}}\\
    \adjustbox{max width=0.98\textwidth}{
    \begin{tabular}{lcccccccc}
    \toprule
    \emph{AE of ATT} & \multicolumn{4}{c}{within sample} & \multicolumn{4}{c}{out-of-sample} \\
    \cmidrule(lr){2-5}\cmidrule(lr){6-9}
    \textsc{Method} & Euclidian & Wasserstein & Hellinger & PS & Euclidian & Wasserstein & Hellinger & PS \\
    \midrule
    VAE & $\underline{0.085 \sd{\pm0.004}}$ & $\underline{0.085 \sd{\pm0.004}}$ &$0.112 \sd{\pm0.006}$ & $0.116 \sd{\pm0.005}$ &  $\underline{0.266 \sd{\pm0.012}}$ & $\underline{0.266 \sd{\pm0.012}}$ &$0.388 \sd{\pm0.021}$ & $0.315 \sd{\pm0.015}$\\
    CEVAE &  $\underline{0.052 \sd{\pm0.002}}$ & $\underline{0.052 \sd{\pm0.002}}$ &$0.109 \sd{\pm0.006}$ & $0.086 \sd{\pm0.004}$ & $\underline{0.174 \sd{\pm0.009}}$ &$\underline{0.174 \sd{\pm0.009}}$ &$0.443 \sd{\pm0.025}$ & $0.255 \sd{\pm0.011}$\\
    TEDVAE &$\underline{0.052 \sd{\pm0.003}}$  &$0.053 \sd{\pm0.003}$ & $0.168 \sd{\pm0.012}$ &$0.087 \sd{\pm0.004}$ & $\underline{0.179 \sd{\pm0.009}}$ & $\underline{0.179 \sd{\pm0.009}}$  & $0.495 \sd{\pm0.028}$ & $0.258 \sd{\pm0.015}$\\
    Ours & $\underline{0.037 \sd{\pm0.002}}$ & $0.040 \sd{\pm0.003}$& $\underline{0.037 \sd{\pm0.003}}$& ${0.085 \sd{\pm0.004}}$&$0.113 \sd{\pm0.007}$&$\underline{0.112 \sd{\pm0.007}}$ & $0.128 \sd{\pm0.008}$ &$0.177 \sd{\pm0.015}$\\
    \quad+I & $\underline{0.038 \sd{\pm0.002}}$& $0.039 \sd{\pm0.002}$ & $0.040 \sd{\pm0.002}$ & $0.079 \sd{\pm0.004}$ &$\underline{0.122 \sd{\pm0.007}}$&$0.124 \sd{\pm0.007}$ &$0.119 \sd{\pm0.007}$ & $0.182 \sd{\pm0.013}$\\
    \quad+sep &${0.037 \sd{\pm0.002}}$ & $\underline{0.036 \sd{\pm0.002}}$ & $0.039 \sd{\pm0.002}$& $0.079 \sd{\pm0.004}$ &$\underline{0.111 \sd{\pm0.007}}$& $0.115 \sd{\pm0.007}$ &${0.116 \sd{\pm0.007}}$  &${0.167 \sd{\pm0.012}}$\\
    \bottomrule
    \end{tabular}}\\
    {\tiny{\emph{mean $\pm$ standard error over 300 random replications; lowest mean \underline{underlined}}}}\\
    \emph{(ii) ATT estimation for \emph{subset + low}}\\
    \adjustbox{max width=0.98\textwidth}{
    \begin{tabular}{lcccccccc}
    \toprule
    \emph{AE of ATT} & \multicolumn{4}{c}{within sample} & \multicolumn{4}{c}{out-of-sample} \\
    \cmidrule(lr){2-5}\cmidrule(lr){6-9}
    \textsc{Method} & Euclidian & Wasserstein & Hellinger & PS & Euclidian & Wasserstein & Hellinger & PS \\
    \midrule
    VAE & $\underline{0.264 \sd{\pm0.014}}$ &$\underline{0.264 \sd{\pm0.014}}$ &$0.288 \sd{\pm0.015}$ & $0.579 \sd{\pm0.071}$  &$\underline{0.427 \sd{\pm0.022}}$ &$\underline{0.427 \sd{\pm0.022}}$ &$0.534 \sd{\pm0.034}$ &$0.697 \sd{\pm0.055}$\\
    CEVAE & $\underline{0.150 \sd{\pm0.009}}$ &$\underline{0.150 \sd{\pm0.009}}$ &$0.277 \sd{\pm0.020}$ &$0.554 \sd{\pm0.072}$ &$0.259 \sd{\pm0.015}$ &$\underline{0.258 \sd{\pm0.015}}$ &$0.539 \sd{\pm0.040}$ &$0.582 \sd{\pm0.045}$\\
    TEDVAE &$\underline{0.144 \sd{\pm0.009}}$ &$\underline{0.144 \sd{\pm0.009}}$ &$0.277 \sd{\pm0.022}$  & $0.548 \sd{\pm0.055}$& $0.272 \sd{\pm0.016}$ &$\underline{0.271 \sd{\pm0.016}}$ &$0.574 \sd{\pm0.048}$ &$0.624 \sd{\pm0.052}$  \\
    Ours &  $0.114 \sd{\pm0.008}$ &$\underline{0.113 \sd{\pm0.008}}$ &$0.120 \sd{\pm0.008}$ & $0.271 \sd{\pm0.023}$&$\underline{0.190 \sd{\pm0.016}}$ &$0.200 \sd{\pm0.017}$ &$0.205 \sd{\pm0.016}$ & $0.252 \sd{\pm0.019}$\\
    \quad+I & $0.110 \sd{\pm0.007}$ &$\underline{0.106 \sd{\pm0.007}}$ &$0.108 \sd{\pm0.007}$ & $0.260 \sd{\pm0.025}$& $\underline{0.176 \sd{\pm0.014}}$ &$0.177 \sd{\pm0.014}$ &$0.179 \sd{\pm0.015}$ & $0.279 \sd{\pm0.029}$ \\
    \quad+sep  & $\underline{0.144 \sd{\pm0.009}}$ & $\underline{0.144 \sd{\pm0.009}}$ &$0.277 \sd{\pm0.022}$ & $0.310 \sd{\pm0.039}$ & $0.272 \sd{\pm0.016}$ &$\underline{0.271 \sd{\pm0.016}}$ &$0.574 \sd{\pm0.048}$ & $0.275 \sd{\pm0.028}$ \\
    \bottomrule
    \end{tabular}}\\
    {\tiny{\emph{mean $\pm$ standard error over 300 random replications; lowest mean \underline{underlined}}}}
\end{table}

\end{document}